\documentclass[twoside,twocolumn,9pt]{article}
\usepackage{extsizes}
\usepackage[super,sort&compress,comma]{natbib} 
\usepackage[version=3]{mhchem}
\usepackage[left=1.5cm, right=1.5cm, top=1.785cm, bottom=2.0cm]{geometry}
\usepackage{balance}
\usepackage{widetext}
\usepackage{times,mathptmx}
\usepackage{sectsty}
\usepackage{graphicx} 
\usepackage{lastpage}
\usepackage[format=plain,justification=raggedright,singlelinecheck=false,font={stretch=1.125,small,sf},labelfont=bf,labelsep=space]{caption}
\usepackage{float}
\usepackage{fnpos}
\usepackage[english]{babel}
\usepackage{array}
\usepackage{droidsans}
\usepackage{charter}
\usepackage[T1]{fontenc}
\usepackage[usenames,dvipsnames]{xcolor}
\usepackage{setspace}
\usepackage[compact]{titlesec}
\usepackage{epstopdf}

\usepackage{amssymb}    
\usepackage{hyperref}   
\usepackage{multirow}   
\usepackage{pifont}     

\newcommand{\bs}[1] {  \boldsymbol{#1}           }
\newcommand{\ff}[1] {  \mbox{\footnotesize{#1}}  }

\newcommand{\xmark}{\ding{55}}%
\newcommand{\cmark}{\ding{51}}%
\DeclareMathOperator*{\argmax}{arg\,max} 

\begin{document}

\thispagestyle{plain}

\makeFNbottom
\makeatletter
\renewcommand\LARGE{\@setfontsize\LARGE{15pt}{17}}
\renewcommand\Large{\@setfontsize\Large{12pt}{14}}
\renewcommand\large{\@setfontsize\large{10pt}{12}}
\renewcommand\footnotesize{\@setfontsize\footnotesize{7pt}{10}}
\makeatother

\renewcommand{\thefootnote}{\fnsymbol{footnote}}
\setcounter{secnumdepth}{5}

\makeatletter 
\renewcommand\@biblabel[1]{#1}            
\renewcommand\@makefntext[1]%
{\noindent\makebox[0pt][r]{\@thefnmark\,}#1}
\makeatother 
\renewcommand{\figurename}{\small{Fig.}~}
\sectionfont{\sffamily\Large}
\subsectionfont{\normalsize}
\subsubsectionfont{\bf}
\setstretch{1.125} 
\setlength{\skip\footins}{0.8cm}
\setlength{\footnotesep}{0.25cm}
\setlength{\jot}{10pt}
\titlespacing*{\section}{0pt}{4pt}{4pt}
\titlespacing*{\subsection}{0pt}{15pt}{1pt}


\setlength{\arrayrulewidth}{1pt}
\setlength{\columnsep}{6.5mm}
\setlength\bibsep{1pt}

\makeatletter 
\newlength{\figrulesep} 
\setlength{\figrulesep}{0.5\textfloatsep}

\makeatother

\twocolumn[
  \begin{@twocolumnfalse}
\sffamily
\begin{tabular}{ p{16cm} }

\noindent\LARGE{\textbf{Deep learning for molecular design\textemdash a review of the state of the art}} \\
\vspace{0.1cm}\\

\noindent\large{Daniel C.\ Elton,\textit{$^{a}$}$^{\dag}$ Zois Boukouvalas,\textit{$^{ab}$} Mark D.\ Fuge,\textit{$^{a}$} and Peter W.\ Chung\textit{$^{a}$}}\\
\vspace{0.1cm}\\
\noindent\normalsize{In the space of only a few years, deep generative modeling has revolutionized how we think of artificial creativity, yielding autonomous systems which produce original images, music, and text. Inspired by these successes, researchers are now applying deep generative modeling techniques to the generation and optimization of molecules\textemdash in our review we found 45 papers on the subject published in the past two years. These works point to a future where such systems will be used to generate lead molecules, greatly reducing resources spent downstream synthesizing and characterizing bad leads in the lab. In this review we survey the increasingly complex landscape of models and representation schemes that have been proposed. The four classes of techniques we describe are recursive neural networks, autoencoders, generative adversarial networks, and reinforcement learning. After first discussing some of the mathematical fundamentals of each technique, we draw high level connections and comparisons with other techniques and expose the pros and cons of each. Several important high level themes emerge as a result of this work, including the shift away from the SMILES string representation of molecules towards more sophisticated representations such as graph grammars and 3D representations, the importance of reward function design, the need for better standards for benchmarking and testing, and the benefits of adversarial training and reinforcement learning over maximum likelihood based training.  } \\
\vspace{0.1cm}\\

\end{tabular}

 \end{@twocolumnfalse} \vspace{0.6cm}
]

\renewcommand*\rmdefault{bch}\normalfont\upshape
\rmfamily
\section*{}
\vspace{-1cm}

\footnotetext{\textit{$^{a}$}~Department of Mechanical Engineering, University of Maryland, College Park, Maryland, 20740, United States of America. E-mail:daniel.elton@nih.gov}
\footnotetext{\textit{$^{b}$}~Department of Mathematics and Statistics, American University, Washington, D.C., 20016, United States of America.}
\footnotetext{\textit{$^{\dag}$}~Present address: National Institutes of Health Clinical Center, Bethesda, Maryland, United States of America.}




The average cost to bring a new drug to market is now well over one billion USD,\cite{DiMasi2016Innovation} with an average time from discovery to market of 13 years.\cite{Paul2010HowToImprove} Outside of pharmaceuticals the average time from discovery to commercial production can be even longer, for instance for energetic molecules it is 25 years.\cite{Homburg2017Remarks} A critical first step in molecular discovery is generating a pool of candidates for computational study or synthesis and characterization. This is a daunting task because the space of possible molecules is enormous\textemdash the number of potential drug-like compounds has been estimated to be between $10^{23}$ and $10^{60}$,\cite{Polishchuk2013:675} while the number of all compounds that have been synthesized is on the order of $10^8$. Heuristics, such as Lipinski's ``rule of five'' for pharmaceuticals\cite{Lipinski1997:3} can help narrow the space of possibilities, but the task remains daunting. High throughput screening (HTS)\cite{Macarron2011HTSreview} and high throughput virtual screening (HTVS)\cite{PyzerKnapp2015HTVS} techniques have made larger parts of chemical space accessible to computational and experimental study. Machine learning has been shown to be capable of yielding rapid and accurate property predictions for many properties of interest and is being integrated into screening pipelines, since it is orders of magnitude faster than traditional computational chemistry methods.\cite{Butler2018MLforMolecules} Techniques for the interpretation and ``inversion'' of a machine learning model can illuminate structure-property relations that have been learned by the model which can in turn be used to guide the design of new lead molecules.\cite{Raccuglia2016Nature, Barnes2018arxiv} However even with these new techniques bad leads still waste limited supercomputer and laboratory resources, so minimizing the number of bad leads generated at the start of the pipeline remains a key priority. The focus of this review is on the use of deep learning techniques for the targeted generation of molecules and guided exploration of chemical space. We note that machine learning (and more broadly artificial intelligence) is having an impact on accelerating other parts of the chemical discovery pipeline as well, via machine learning accelerated ab-initio simulation,\cite{Butler2018MLforMolecules} machine learning based reaction prediction,\cite{Fooshee2018DeepLearning,Schwaller2018} deep learning based synthesis planning,\cite{Segler2018Planning} and the development of high-throughput ``self-driving'' robotic laboratories.\cite{Henson2018ACSrobots,Roch2018ChemOS}

Deep neural networks, which are often defined as networks with more than three layers, have been around for many decades but until recently were difficult to train and fell behind other techniques for classification and regression. By most accounts, the deep learning revolution in machine learning began in 2012, when deep neural network based models began to win several different competitions for the first time. First came a demonstration by Cire{\c{s}}an et al.\ of how deep neural networks could achieve near-human performance on the task of  handwritten digit classification.\cite{Ciregan2012CVPR} Next came groundbreaking work by Krizhevsky et al.\ which showed how deep convolutional networks achieved superior performance on the 2010 ImageNet image classification challenge.\cite{Krizhevsky2012NIPS} Finally, around the same time in 2012, a multitask neural network developed by Dahl et al.\ won the ``Merck Molecular Activity Challenge" to predict the molecular activities of molecules at 15 different sites in the body, beating out more traditional machine learning approaches such as boosted decision trees.\cite{Dahl2014:arXiv} One of the key technical advances published that year and used by both Krizhevsky et al.\ and Dahl et al.\ was a novel regularization trick called ``dropout''.\cite{Hinton2012Dropout} Another important technical advance was the efficient implementation of neural network training on graphics processing units (GPUs). By 2015 better hardware, deeper networks, and a variety of further technical advances had reduced error rates on the ImageNet challenge by a factor of 3 compared to the Krizhevsky's 2012 result.\cite{Krizhevsky2017:ImageNet}

In addition to the tasks of classification and regression, deep neural networks began to be used for generation of images, audio, and text, giving birth to the field of ``deep generative modeling''. Two key technical advances in deep generative modeling were the variational autoencoder (Kimga et al., 2013\cite{Kingma2013:arxiv}) and generative adversarial networks (Goodfellow et al.\ 2014\cite{Goodfellow2014:2672}). The first work demonstrating deep generative modeling of molecules was the ``molecular autoencoder'' work of G\'{o}mez-Bombarelli et al.\ which appeared on the arXiv in October 2016 and was published in \textit{ACS Central Science} in 2018.\cite{GmezBombarelli2018:268} Since then, there has been an explosion of advancements in deep generative modeling of molecules using several different deep learning architectures and many variations thereof, as shown in table~\ref{generativemodelingpapers}. In addition to new architectures, new representation schemes, many of which are graph based, have been introduced as alternatives to the SMILES representation used by G\'{o}mez-Bombarelli et al. The growing complexity of the landscape of architectures and representations and the lack of agreement upon standards for benchmarking and comparing different approaches has prompted us to write this review. 

While much of the work so far has focused on deep generative modeling for drug molecules,\cite{Griffen2018Canwe} there are many other application domains which are benefiting from the application of deep learning to lead generation and screening, such as organic light emitting diodes,\cite{GmezBombarelli2016OLED} organic solar cells,\cite{Jorgensen2018JCP} energetic materials,\cite{Elton2018scirep,Barnes2018arxiv} electrochromic devices,\cite{Rinderspacher2018Enriched} polymers,\cite{Li2018Tuning} polypeptides,\cite{Nagarajan2017polypeptide,Muller2018polypeptide,Grisoni2018polypeptide} and metal organic frameworks.\cite{Shen2018manifoldMOFs,He2018MOFs} 

Our review touches on four major issues we have observed in the field. The first is the importance and opportunities for improvement by using different molecular representations. Recent efforts have begun to depart from the use of Simplified Molecular-Input Line-Entry System (SMILES) strings towards representations that are ``closer to the chemical structure'' and offer improved chemical accuracy, such as graph grammar based methods. The second issue is architecture selection.  We discuss the pros and cons underlying different choices of model architecture and present some of their key mathematical details to better illuminate how different approaches relate to each other.  This leads us to highlight the advantages of adversarial training and reinforcement learning over maximum likelihood based training. We also touch on techniques for molecular optimization using generative models, which has grown in popularity recently. The third major issue is the approaches for quantitatively evaluating different approaches for molecular generation and optimization. Fourth, and finally, we discuss is reward function design, which is crucial for the practical application of methods which use reinforcement learning. We contribute by offering novel overview of how to engineer reward function to generate a set of leads which is chemically stable, diverse, novel, has good properties, and is synthesizable.  

There are reasons to be skeptical about whether today's deep generative models can outperform traditional computational approaches to lead generation and optimization. Traditional approaches are fundamentally combinatorial in nature and involve mixing scaffolds, functional groups, and fragments known to be relevant to the problem at hand (for a review, see Pirard et al.\cite{Pirard2011:255}). A naive combinatorial approach to molecular generation leads to most molecules being unstable or impossible to synthesize, so details about chemical bonding generally must be incorporated. One approach is to have an algorithm perform virtual chemical reactions, either from a list of known reactions, or using ab initio methods for reaction prediction.\cite{Wang2014Discovering} Another popular approach is to use genetic algorithms with custom transformation rules which are known to maintain chemical stability.\cite{Besnard2012:215} One of the latest genetic algorithm based approaches (``Grammatical Evolution'') can match the performance of the deep learning approaches for molecular optimization under some metrics.\cite{Yoshikawa2018Population} Deep generative modeling of molecules has made rapid progress in just a few years and there are reasons to expect this progress to continue, not just with better hardware and data, but due to new architectures and approaches. For instance, generative adversarial networks and deep reinforcement learning (which may be combined or used separately) have both seen technical advancements recently. 

\tableofcontents

\section{Molecular representation}
The molecular representation refers to the digital encoding used for each molecule that serves as input for training the deep learning model. A representation scheme must capture essential structural information about each molecule. Creating an appropriate representation from a molecular structure is called featurization. Two important properties that are desirable (but not required) for representations are uniqueness and invertibility. Uniqueness means that each molecular structure is associated with a single representation. Invertibility means that each representation is associated with a single molecule (a one-to-one mapping). Most representations used for molecular generation are invertible, but many are non-unique. There are many reasons for non-uniqueness, including the representation not being invariant to the underlying physical symmetries of rotation, translation, and permutation of atomic indexes. 

Another factor one should consider when choosing a representation is the whether it is a character sequence or tensor. Some methods only work with sequences, while others only work with tensor. Sequences may be converted into tensors using one-hot encoding. Another choice is whether to use a representation based on the 3D coordinates of the molecule or a representation based on the 2D connectivity graph. Molecules are fundamentally three dimensional quantum mechanical objects, typically visualized as consisting of nuclei with well-defined positions surrounded by many electrons which are described by a complex-valued wavefunction. Fundamentally, all properties of a molecule can be predicted using quantum mechanics given only the relative coordinates of the nuclei and the type and ionization state of each atom. However, for many applications, working with 3D representations is cumbersome and unnecessary. In this section, we review both 3D and 2D representation schemes that have been developed recently. 
\begin{table}[h]
  \small
  \centering
  \caption{\ Different representation schemes}
  \label{tab:representations}
    \begin{tabular}{c| c c c }
    & method      &   unique? & invertible?  \\
    \hline
\multirow{3}{*}{\rotatebox{90}{3D }}  
  & raw voxels            &  \xmark &  \cmark  \\
  & smoothed voxels        &  \xmark &  \cmark \\
  & tensor field networks &  \xmark & \xmark \\
\hline 
\multirow{6}{*}{\rotatebox{90}{2D graph}} 
 & SMILES             & \xmark   & \cmark \\
 & canonical SMILES   & \cmark   & \cmark \\
 & InChI              & \cmark   & \cmark \\
 & MACCS keys         & \cmark   & \xmark \\
 & tensors            & \xmark   & \cmark \\
 & Chemception images & \cmark   & \cmark \\
 & fingerprinting     & \cmark   & \xmark \\
    \end{tabular}
\end{table} 

\subsection{Representations of 3D geometry}
Trying to implement machine learning directly with nuclear coordinates introduces a number of issues. The main issue is that coordinates are not invariant to molecular translation, rotation, and permutation of atomic indexing. While machine learning directly on coordinates is possible, it is much better to remove invariances to create a more compact representation (by removing degrees of freedom) and develop a scheme to obtain a unique representation for each molecule. One approach that uses 3D coordinates uses a 3D grid of voxels and specifies the nuclear charge contained within each voxel, thus creating a consistent representation. Nuclear charge (i.e.\ atom type) is typically specified by a one-hot vector of dimension equal to the number of atom types in the dataset. This scheme leads to a very high dimensional sparse representation, since the vast majority of voxels will not contain a nuclear charge. While sparse representations are considered desirable in some contexts, here sparsity leads to very large training datasets. This issue can be mitigated via spatial smoothing (blurring) by placing spherical Gaussians or a set of decaying concentric waves around each atomic nuclei.\cite{Kuzminykh2018asap} Alternatively, the van der Waals radius may be used.\cite{Skalic2019asap} Amidi et al.\ use this type of approach for predictive modeling with 3D convolutional neural networks (CNNs),\cite{Amidi2018enzynet} while Kuzminykh et al.\ and Skalic et al.\ use this approach with a CNN-based autoencoder for generative modeling.\cite{Kuzminykh2018asap,Skalic2019asap} 

Besides high dimensionality and sparsity, another issue with 3D voxelized representations is they do not capture invariances to translation, rotation, and reflection, which are hard for present-day deep learning based architectures to learn. Capturing such invariances is important for property prediction, since properties are invariant to such transformations. It is also important for creating compact representations of molecules for generative modeling. One way to deal with such issues is to always align the molecular structure along a principal axis as determined by principle component analysis to ensure a unique orientation.\cite{Amidi2018enzynet,Kuzminykh2018asap} Approaches which generate feature vectors from 3D coordinates that are invariant to translation and rotation are wavelet transform invariants,\cite{Hirn2017wavelet} solid harmonic wavelet scattering transforms,\cite{Eickenberg2017:6540} and tensor field networks.\cite{thomas2018arXivtensor} All of these methods incur a loss of information about 3D structure and are not easy to invert, so their utility for generative modeling may be limited (deep learning models learn to generate these representations, but if they cannot be unambiguously related to a 3D structure they are not very useful). Despite their issues with invertibility, tensor field networks have been suggested to have utility for generative modeling since it was shown they can accurately predict the location of missing atoms in molecules where one atom was removed.\cite{thomas2018arXivtensor} We expect future work on 3D may proceed in the direction of developing invertible representations that are based on the internal (relative) coordinates of the molecule. 

\subsection{Representations of molecular graphs} 
\subsubsection{SMILES and string-based representations}
Most generative modeling so far has not been done with coordinates but instead has worked with molecular graphs. A molecule can be considered as an undirected graph $\mathcal{G}$ with a set of edges $\mathcal{E}$ and set of vertices $\mathcal{V}$. The obvious disadvantage of such graphs is that information about bond lengths and 3D conformation is lost. For some properties one may wish to predict, the specific details of a molecule's 3D conformations may be important. For instance, when packing in a crystal or binding to a receptor, molecules will find the most energetically favorable conformation, and details of geometry often have a big effect. Despite this, graph representations have been remarkably successful for a variety of generative modeling and property prediction tasks. If a 3D structure is desired from a graph representation, molecular graphs can be embedded in 3D using distance geometry methods (for instance as implemented in the \textit{OpenBabel} toolkit\cite{OBoyle2011openbabel,OpenBabel}). After coordinate embedding, the most energetically favorable conformation of the molecule can be obtained by doing energy minimization with classical forcefields or quantum mechanical simulation.

There are several ways to represent graphs for machine learning. The most popular way is the SMILES string representation.\cite{Weininger1988} SMILES strings are a non-unique representation which encode the molecular graph into a sequence of ASCII characters using a depth-first graph traversal. SMILES are typically first converted into a one-hot based representation. Generative models then produce a categorical distribution for each element, often with a softmax function, which is sampled. Since standard multinomial sampling procedures are non-differentiable, sampling can be avoided during training or a Gumbel-softmax can be used.\cite{Jang2016arXivGumbel,deCao2018molganICML}


Many deep generative modeling techniques have been developed specifically for sequence generation, most notably Recurrent Neural Networks (RNNs), which can be used for SMILES generation. The non-uniqueness of SMILES arises from a fundamental ambiguity about which atom to start the SMILES string construction on, which means that every molecule with $N$ heavy (non-hydrogen) atoms can have at least $N$ equivalent SMILES string representations. There is additional non-uniqueness due to different conventions on whether to include charge information in resonance structures such as nitro groups and azides. The \textit{MolVS}\cite{MolVS} or \textit{RDKit}\cite{rdkit} cheminformatics packages can be used to standardize SMILES, putting them in a canonical form. However, Bjerrum et al.\ have pointed out that the latent representations obtained from canonical SMILES may be less useful because they become more related to specific grammar rules of canonical SMILES rather than the chemical structure of the underlying molecule.\cite{Bjerrum2018arxivLatent} This is considered an issue for interpretation and optimization since it is better if latent spaces encode underlying chemical properties and capture notions of chemical similarity rather than SMILES syntax rules. Bjerrum et al.\ have suggested SMILES enumeration (training on all SMILES representations of each molecule), rather than using canonical SMILES, as a better solution to the non-uniqueness issue.\cite{Bjerrum2018arxivLatent,Bjerrum2017arxivenumeration} An approach similar to SMILES enumeration is used in computer vision applications to obtain rotational invariance\textemdash image datasets are often ``augmented'' by including many rotated versions of each image. Another approach to obtain better latent representations explored by Bjerrum et al.\ is to input both enumerated SMILES and Chemception-like image arrays (discussed below) into a single ``heteroencoder'' framework.\cite{Bjerrum2018arxivLatent}

In addition to SMILES strings, G\'{o}mez-Bombarelli et al.\ have tried InChI strings\cite{Heller2013:7} with their variational autoencoder, but found they led to inferior performance in terms of the decoding rate and the subjective appearance of the molecules generated. Interestingly, Winter et al.\ show that more physically meaningful latent spaces can be obtained by training a variational autoencoder to translate between InChI to SMILES.\cite{Winter2019Learning} There is an intuitive explanation for this\textemdash the model must learn to extract the underlying chemical structures which are encoded in different ways by the two representations. 

SMILES based methods often struggle to achieve a high percentage of valid SMILES. As a possible solution to this, Kusner et al.\ proposed decomposing SMILES into a sequence of rules from a context free grammar (CFG).\cite{KusnerArxiv2017} The rules of the context-free grammar impose constraints based on the grammar of SMILES strings.\cite{Dai2018arxivSDVAE} Because the construction of SMILES remains probabilistic, the rate of valid SMILES generation remains below 100\%, even when CFGs are employed and additional semantic constraints are added on top.\cite{Dai2018arxivSDVAE} Despite the issues inherent with using SMILES, we expect it will continue to a popular representation since most datasets store molecular graphs using SMILES as the native format, and since architectures developed for sequence generation (i.e.\ for natural language or music) can be readily adopted. Looking longer term, we expect a shift towards methods which work directly with the graph and construct molecules according to elementary operations which maintain chemical valence rules. 

Li et al.\ have developed a conditional graph generation procedure which obtains a very high rate of valid chemical graphs (91\%) but lower negative log likelihood scores compared to a traditional SMILES based RNN model.\cite{Li2018arxivLearning} Another more recent work by Li et al.\ uses a deep neural network to decide on graph generation steps (append, connect, or terminate).\cite{Li2018asap} Efficient algorithms for graph and tree enumeration have been previously developed in a more pure computer science context. Recent work has looked at how such techniques can be used for molecular graph generation,\cite{Suzuki2014} and likely will have utility for deep generative models as well. 

\subsubsection{Image-based representations}
Most small molecules are easily represented as 2D images (with some notable exceptions like cubane). Inspired by the success of Google's Inception-ResNet deep convolutional neural network (CNN) architecture for image recognition, Goh et al.\ developed ``Chemception'', a deep CNN which predicts molecular properties using custom-generated images of the molecular graph.\cite{Goh2017arxiv} The Chemception framework takes a SMILES string in and produces an 80x80 greyscale image which is actually an array of integers, where empty space is `0', bonds are `2' and atoms are represented by their atomic number.\cite{Goh2017arxiv} Bjerrum et al.\ extend this idea, producing ``images'' with five color channels which encode a variety of molecular features, some which have been compressed to few dimensions using PCA.\cite{Bjerrum2018arxivLatent}

\subsubsection{Tensor representations}
Another approach to storing the molecular graph is to store the vertex type (atom type), edge type (bond type), and connectivity information in multidimensional arrays (tensors).  In the approach used by de Cao \& Kipf,\cite{deCao2018molganICML,deCao2018:arxiv} each atom is a vertex $v_i \in \mathcal{V}$ which may be represented by a one-hot vector $\bs{x}_i \in \lbrace 0,1 \rbrace^{|\mathcal{A}|}$ which indicates the atom type, out of $|\mathcal{A}|$ possible atom types. Each bond is represented by an edge $(v_i, v_j)$ which is associated with a one-hot vector $\bs{y}_i \in \lbrace 0,1 \rbrace^Y$ representing the type of bond out of $Y$ possible bond types. The vertex and edge information can be stored in a vertex feature matrix $X = [\bs{x}_1, \dots, \bs{x}_N]^T \in \mathbb{R}^{N\times |\mathcal{A}|}$ and an adjacency tensor $A \in \mathbb{R}^{N\times N\times Y}$ where $A_{ij} \in \mathbb{R}^Y $. Simonovsky et al.\cite{Simonovsk2018arxiv} use a similar approach\textemdash they take a vertex feature matrix $X$ and concatenate the adjacency tensor $A$ with a traditional adjacency matrix where connections are indicated by a `1'. As with SMILES, adjacency matrices suffer from non-uniqueness\textemdash for a molecule with $N$ atoms, there are $N!$ equivalent adjacency matrices representing the same molecular graph, each corresponding to a different re-ordering of the atoms/nodes. This makes it challenging to compute objective functions, which require checking if two adjacency matrix representations correspond to the same underlying graph (the ``graph isomorphism '' problem, which takes $N^4$ operations in the worse case). Simonovsky et al.\ use an approximate graph matching algorithm to do this, but it is still computationally expensive. 

\subsubsection{Other graph-based representations}
Another approach is to train an RNN or reinforcement learning agent to operate directly on the molecular graph, adding new atoms and bonds in each action step from a list of predefined possible actions. This approach is taken with the graph convolutional policy network\cite{You2018arXiv} and in recent work using pure deep reinforcement learning to generate molecules.\cite{Zhou2018arXiv} Because these methods work directly on molecular graphs with rules which ensure that basic atom valence is satisfied, they generate 100\% chemically valid molecules. 

Finally, when limited to small datasets one may elect to do generative modeling with compact feature vectors based on fingerprinting methods or descriptors. There are many choices (Coulomb matrices, bag of bonds, sum over bonds, descriptor sets, graph convolutional fingerprints, etc.) which we have previously tested for regression,\cite{Elton2018scirep,Boukouvalas2018independentarxiv} but they are generally not invertible unless a very large database with a look-up table has been constructed. (In this context, by invertible we mean the complete molecular graph can be reconstructed without loss.) As an example of how it may be done, Kadurin et al.\ use 166 bit Molecular ACCess System (MACCS) keys\cite{Durant2002:1273} for molecular representation with adversarial autoencoders.\cite{Kadurin2016AAE, Kadurin2017:3098} In MACCS keys, also called MACCS fingerprints, each bit is associated with a specific structural pattern or question about structure. To associate molecules to MACCS keys one must search for molecules with similar or identical MACCS keys in a large chemical database. Fortunately several large online chemical databases have application programming interfaces (APIs) which allow for MACCS-based queries, for instance \textit{PubChem}, which contains 72 million compounds. 

\begin{figure*}[h]
  \centering
  \includegraphics[width=18cm]{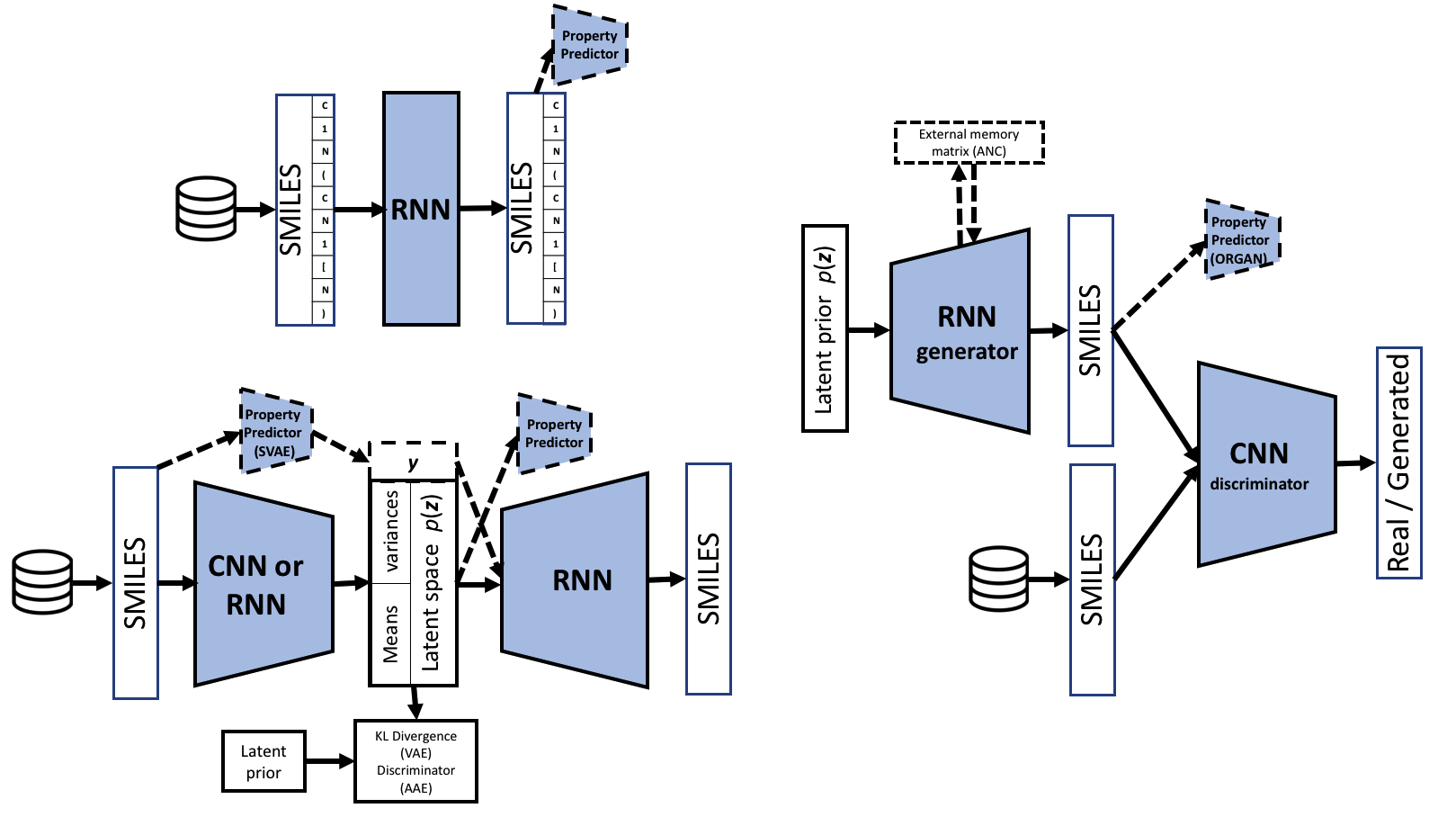}
  \caption{ Bird's eye views of three popular frameworks for generative modeling using SMILES strings, with possible variations shown with dashed lines}
  \label{fig:archdiagram}
\end{figure*}

\begin{table*}
    \small
    \centering
    \caption{ For works that trained models separately on multiple datasets we report only the largest dataset used. Several of these datasets are described in table \ref{datasetslist}, which lists the major publicly available datasets. Other datasets are ``HCEP'', the Harvard Clean Energy Project dataset of lead molecules for organic photovoltaic,  ``PSC'', a dataset of monomer repeat units for polymer solar cells, ``MCF-7'', a database of anti-cancer molecules, and ``L1000'', a database of molecules and gene expression profiles.\\ 
    Acronyms used are: AAE = adversarial autoencoder, ANC = adversarial neural computer, ATNC = adversarial threshold neural computer, BMI = Bayesian model inversion, CAAE = constrained AAE,  CCM-AAE = Constant-curvature Riemannian manifold AAE,  CFG = context free grammar, CVAE = constrained VAE, ECC = edge-conditioned graph convolutions,\cite{Simonovsky2017:29CVPR}GAN = generative adversarial network, GCPN = graph convolutional policy network, GVAE = grammar VAE,  JT-VAE = junction tree VAE,  MHG = molecular hypergraph grammar, MPNN = Message Passing Neural Net, RG = reduced graph, RNN = recurrent neural network, sGAN = stacked GAN, SD-VAE = syntax-directed VAE, SSVAE = semi-supervised VAE, VAE = variational autoencoder \\
    $^*$ filtered to isolate likely bioactive compounds. 
    }
    \label{generativemodelingpapers}
    \begin{tabular}{c c r r r}
architecture & representation & $N_{\ff{train}}$ & dataset(s) & citation(s) \\ 
\hline
 RNN    & SMILES          &  1,611,889    & ZINC   & Bjerrum, 2017 \cite{Bjerrum2017arxiv} \\
 RNN    & SMILES          &  541,555      & ChEMBL$^*$ & Gupta, 2017  \cite{Gupta2017:1700111} \\ 
 RNN    & SMILES          &   350,419     & DRD2   & Oliverona, 2017 \cite{Olivecrona2017} \\
 RNN    & SMILES          &  1,400,000    & ChEMBL & Segler, 2017 \cite{Segler2017:120} \\ 
 RNN    & SMILES          & 250,000       & ZINC   & Yang, 2017 \cite{Yang2017:972} \\
 RNN    & SMILES          & 200,000       & ZINC   & Cherti, 2017 \cite{cherti2017novo} \\  
 RNN    & SMILES          &  1,735,442    & ChEMBL & Neil, 2018 \cite{neil2018exploring} \\ 
 RNN    & SMILES          & 1,500,000     & ChEMBL & Popova, 2018 \cite{Popova2018} \\
 RNN    & SMILES          & 13,000        & PubChemQC & Sumita, 2018 \cite{SumitapreprintACScentral} \\ 
 RNN    & SMILES          & 541,555       & ChEMBL$^*$ & Merk, 2018 \cite{Merk2018:1700153} \\
 RNN    & SMILES          & 541,555       & ChEMBL$^*$ & Merk, 2018 \cite{Merk2018Tuning}\\
 RNN    & SMILES          & 509,000       & ChEMBL & Ertl, 2018 \cite{Ertl2017:arxiv} \\
 RNN    & SMILES          & 1,000,000     & GDB-13 & Ar\'{u}s-Pous, 2018 \cite{Arus-Pous2018Exploring}\\
 RNN    & SMILES         & 163,000       & ZINC   & Zheng, 2019 \cite{Zheng2019QBMG}\\
 RNN    & RG+SMILES       & 798,243       & ChEMBL & Pog\'{a}ny, 2018 \cite{Pogny2018DeNovo}\\
 RNN    & graph operations& 130,830       & ChEMBL & Li, 2018 \cite{Li2018arxivLearning}\\
 \hline
 VAE    & SMILES         & 249,000       & ZINC/QM9 & G\'{o}mez-Bombarelli, 2016 \cite{GmezBombarelli2018:268}\\
 VAE    & SMILES         & 1,200,000     & ChEMBL  & Blaschke, 2018 \cite{Blaschke2017:1700123}  \\ 
 VAE    & SMILES         & 500,000       & ZINC    & Lim, 2018 \cite{Lim2018springerASAP}  \\ 
 VAE    & SMILES         & 300,000       & ZINC    & Kang, 2018 \cite{Kang2018}  \\ 
 VAE    & SMILES         & 190,000       & ZINC    & Harel, 2018 \cite{Harel2018prototype}  \\ 
 VAE    & SMILES         & 1,211,352     & ChEMBL23 & Sattarov, 2019 \cite{Sattarov2019DeNovo}  \\ 
 GVAE   & CFG (SMILES)   & 200,000       & ZINC    & Kusner, 2017 \cite{2017KusnerArxiv}  \\ 
 GVAE   & CFG (custom)   & 3,989         & PSC     & J{\o}rgensen, 2018 \cite{Jorgensen2018JCP,Jorgensen2018:1700133}\\ 
SD-VAE  & CFG (custom)   & 250,000       &  ZINC   & Dai, 2018 \cite{Dai2018arxivSDVAE}  \\ 
JT-VAE  & graph operations& 250,000       & ZINC    & Jin, 2018 \cite{Jin2018arxiv}  \\ 
JT-VAE  & graph operations& 250,000       & ZINC    & Jin, 2019 \cite{JinICLR2019}  \\ 
CVAE    & graph          &  250,000      & ZINC/CEPDB & Liu, 2018 \cite{Liu2018arxiv} \\ 
MHG-VAE & graph (MHG)    &  220,011      & ZINC    & Kajino, 2018 \cite{Kajino2018arxivMHGVAE}\\
 VAE    & graph          &  72,000,000   & ZINC+PubChem & Winter, 2018 \cite{Winter2018chemrxiv} \\ 
 VAE    & graph          & 10,000        & ZINC/QM9 & Samanta, 2018 \cite{Samanta2018:arxiv}  \\ 
 VAE    & graph (tensors)&  10,000   & ZINC     & Samanta, 2018 \cite{Samanta2018arXivNaVAE} \\
 VAE    & graph (tensors)&  250,000      & ZINC/QM9 & Simonovsky, 2018 \cite{Simonovsk2018arxiv} \\ 
CVAE    & graph (tensors)&  250,000       & ZINC    & Ma, 2018 \cite{Ma2018Constrained} \\
 VAE    & 3D wave transform &  4,8000,000& ZINC     & Kuzminkykh, 2018 \cite{Kuzminykh2018asap}  \\ 
CVAE    & 3D density     &  192,813,983   & ZINC    & Skalic, 2019 \cite{Skalic2019asap} \\
 VAE+RL &  MPNN+graph ops   &  133,885      &  QM9    & Kearns, 2019\cite{Kearns2019arxiv}  \\
\hline
 GAN    & SMILES          &  5,000        & GBD-17  & Guimaraes, 2017 \cite{Guimaraes2017ORGANarxiv}\\
 GAN (ANC) & SMILES       &    15,000     & \footnotesize{ZINC/CHEMDIV}& Putin, 2018\cite{Putin2018:1194}\\
GAN (ATNC) & SMILES       &    15,000     & \footnotesize{ZINC/CHEMDIV}& Putin, 2018\cite{Putin2018asap}\\
 GAN    & graph (tensors) &  133,885      & QM9     & De Cao, 2018\cite{deCao2018:arxiv,deCao2018molganICML}\\
 GAN    & MACCS (166bit)  & 6,252         & MCF-7   & Kadurin, 2017 \cite{Kadurin2017:3098}\\
 sGAN   & MACCS (166bit)  & 20,000        & L1000   & M\'{e}ndez-Lucio, 2017 \cite{MendezLucio2018chemarxiv}\\
CycleGAN& graph operations& 250,000       & ZINC    & Maziarka, 2019 \cite{maziarka2019molcycleganArxiv}\\
\hline
 AAE    & MACCS (166bit) &  6,252        & MCF-7   & Kadurin, 2017 \cite{Kadurin2016AAE}\\
 AAE    & SMILES         &  15,000       & HCEP    & Sanchez-Lengeling, 2017 \cite{Sanchez-Lengeling2017ORGANICchemrxiv}\\
 CCM-AAE& graph (tensors)    &  133,885    & QM9    & Grattarola, 2018 \cite{Grattarola2018} \\
 BMI    & SMILES          & 16,674       & PubChem & Ikebata, 2017 \cite{ Ikebata2017:379} \\ 
 CAAE   & SMILES         &  1,800,000    & ZINC    & Polykovskiy, 2018 \cite{Polykovskiy2018asap}\\
GCPN    & graph          &  250,000      & ZINC    & You, 2018 \cite{You2018arXiv}\\ 
 pure RL & graph         &  n/a         &  n/a    & Zhou, 2018 \cite{Zhou2018arXiv} \\
 pure RL & fragments     &  n/a         &  n/a    & St\aa hl, 2019 \cite{Stahl2019:chemarxiv} \\
  \end{tabular}
\end{table*}

\begin{table*}
    \centering
    \caption{ Some publicly available datasets. $^*$also contains numerous conformers for each molecule, for a total of 4,855 structures}
    \label{datasetslist}
    \begin{tabular}{c p{7cm} r  p{5cm}}
dataset & description & $N$ & URL / citation \\ 
\hline
GDB-13  & Combinatorially generated library. &  977,468,314 & \href{http://gdb.unibe.ch/downloads/}{http://gdb.unibe.ch/downloads/}\cite{Blum2009GDB13}\\
ZINC15    & Commercially available compounds. & $>$750,000,000 & \href{http://zinc15.docking.org}{http://zinc15.docking.org}\cite{Sterling2015Zinc} \\
GDB-17  & Combinatorially generated library. &  50,000,000 & \href{http://gdb.unibe.ch/downloads/}{http://gdb.unibe.ch/downloads/}\cite{Ruddigkeit2012GDB17}\\
eMolecules  & Commercially available compounds. & 18,000,000 & \href{https://reaxys.emolecules.com/}{https://reaxys.emolecules.com/} \\
SureChEMBL & Compounds obtained from chemical patents. & 17,000,000 & \href{https://www.surechembl.org/search/}{https://www.surechembl.org/search/} \\ 
PubChemQC  & Compounds from PubChem with property data from quantum chemistry (DFT) calculations. & 3,981,230 &  \href{http://pubchemqc.riken.jp/}{http://pubchemqc.riken.jp/}\cite{Nakata2017PubChemQC} \\ 
ChEMBL  & A curated database of bioactive molecules. & 2,000,000 & \href{https://www.ebi.ac.uk/chembl/}{https://www.ebi.ac.uk/chembl/} \\
SuperNatural & A curated database of natural products. & 2,000,000 & \href{http://bioinformatics.charite.de/supernatural/}{http://bioinformatics.charite.de/supernatural/} \\
QM9 & Stable small CHONHF organic molecules taken from GDB-17 with properties calculated from ab initio density functional theory. & 133,885 & \href{http://quantum-machine.org/datasets/}{http://quantum-machine.org/datasets/}\cite{ramakrishnan2014quantum} \\
BNPAH & B, N-substituted polycyclic aromatic hydrocarbons with properties calculated from ab initio density functional theory. & 33,000 & \href{https://moldis.tifrh.res.in/datasets.html}{https://moldis.tifrh.res.in/datasets.html}\cite{Chakraborty2019TheChemicalSpace} \\
DrugBank & FDA drugs and other drugs available internationally. & 10,500 & \href{https://www.drugbank.ca/}{https://www.drugbank.ca/} \\
Energetics & Energetic molecules and simulation data collected from public domain literature. & 417 & \href{https://git.io/energeticmols}{https://git.io/energeticmols}\cite{Elton2018scirep} \\
HOPV15 & Harvard Organic Photovoltaic Dataset & 350$^*$ & \href{https://figshare.com/articles/HOPV15_Dataset/1610063/4}{https://figshare.com/articles
/HOPV15\_Dataset/1610063/4}\cite{Lopez2016TheHarvard} \\
\hline
 \end{tabular}
\end{table*}

\section{Deep learning architectures}
In this section we summarize the mathematical foundations of several popular deep learning architectures and expose some of their pros and cons. A basic familiarity with machine learning concepts is assumed. 

\subsection{Recurrent neural networks (RNNs)}
We discuss recurrent neural network sequence models first because they are fundamental to molecular generation\textemdash most VAE and GAN implementations include an RNN for sequence generation. In what follows, a sequence of length $T$ will be denoted as $S_{1:T} = (s_1, \cdots, s_T), s_t \in \mathcal{V}$, where $\mathcal{V}$ is the set of tokens, also called the vocabulary. For the purpose of this section we assume the sequences in question are SMILES strings, as they are by far the most widely used. As discussed previously in the context of SMILES the  ``tokens'' are the different characters which are used to specify atom types, bond types, parentheses, and the start and stop points of rings. The first step in sequence modeling is typically one-hot encoding of the sequence's tokens, in which each token is represented as a unique $N$ dimensional vector where one element is $1$ and the rest are $0$ (where $N$ is the number of tokens in the vocabulary). 

Recurrent neural networks (RNNs) are the most popular models for sequence modeling and generation. We will not go into detail of their architecture, since it is well described elsewhere.\cite{Geron:2017, GoodfellowBook2016} An important detail to note however is that the type of RNN unit that is typically used for generating molecules is either the long short term memory (LSTM) unit,\cite{Hochreiter1997:1735} or a newer more computationally efficient variant called the gated recurrent unit (GRU).\cite{ChoGRUpaper} Both LSTMs and GRUs contain a memory cell which alleviates the exploding and vanishing gradient problems that can occur when training RNNs to predict long-term dependencies.\cite{Hochreiter1997:1735,ChoGRUpaper}

Sequence models are often trained to predict just a single missing token in a sequence, as trying to predict more than one token leads to a combinatorial explosion of possibilities. Any machine learning model trained to predict the next character in an input sequence can be run in ``generative mode" or ``autoregressive mode" by concatenating the predicted token to the input sequence and feeding the new sequence back into the model. 
However, this type of autoregressive generation scheme typically fails because the model was trained to predict on the data distribution and not its own generative distribution, and therefore each prediction contains at least a small error. As the network is run recursively, these errors rapidly compound, leading to rapid degradation in the quality of the generated sequences. This problem is known as ``exposure bias''.\cite{Ranzato2015arxiv} The Data as Demonstrator (DaD) algorithm tries to solve the problem of exposure bias by running a model recursively during training and comparing the output to the training data during training.\cite{Venkatraman2015:3024} DaD was extended to sequence generation with RNNs by Bengio et al., who called the method ``scheduled sampling".\cite{Bengio2015:1171} While research continues in this direction, issues have been raised about the lack of a firm mathematical foundation for such techniques, with some  suggesting they do not properly approximate maximum likelihood.\cite{Huszard2015:arxiv} 

Better generative models can be obtained by training using maximum likelihood maximization on the sequence space rather than next-character prediction. In maximum likelihood training a model $\pi_\theta$ parametrized by $\theta$ is trained with the following differentiable loss: 
\begin{equation}
    L^{\ff{MLE}} = -\sum\limits_{s \in \mathcal{Z}} \sum\limits_{t=2}^T \log \pi_\theta (s_t  | S_{1:t-1}) 
\end{equation}
Here $\mathcal{Z}$ is the set of training sequences which are assumed to each be of length $T$. This expression is proportional to the negative cross entropy of the model distribution and the training data distribution (maximizing likelihood is equivalent to minimizing cross entropy). MLE training can be done with standard gradient descent techniques and backpropagation through time to compute the gradient of the loss. In practice though this type of training fails to generate valid SMILES, likely because of strict long term dependencies such as closing parentheses and rings. The ``teacher forcing'' training procedure\cite{Williams1989:270} is an important ingredient which was found to be necessary to include in the molecular autoencoder VAE to capture such long term dependencies\textemdash otherwise the generation rate of valid SMILES was found to be near 0\%.\cite{Gomez-BombarelliYouTube} In teacher forcing, instead of sampling from the model's character distribution to get the next character, the right character is given directly to the model during training.\cite{GoodfellowBook2016}

In the context of SMILES strings generation, to generate each character the output layer usually gives probabilities $p_i$ for every possible SMILES string token. When running in generative mode, the typical method is to use a multinomial sampler to sample this distribution, while in training mode one typically just chooses the token with the highest probability. Using a multinomial sampler captures the model's true distribution, but because MLE training tends to focus on optimizing the peaks of the distribution and doesn't always capture the tails of distributions well. So called ``thermal'' rescaling can be used to sample further away from the peaks of the distribution by rescaling the probabilities as: 
\begin{equation}
    p_i^{\ff{new}} =\frac{ \exp\left( \frac{p_i}{T} \right) }{  \sum_i \exp\left( \frac{ p_i }{T} \right)} 
\end{equation}
where $T$ is a sampling temperature. Alternatively, if a softmax layer is used to generate the final output of a neural network, a temperature parameter can be built directly into it. Another alternative is the ``freezing function'': 
\begin{equation}
    p_i^{\ff{new}} =\frac{  p_i^{\frac{1}{T}}  }{  \sum_i p_i^{\frac{1}{T}} } 
\end{equation}
Sampling at low $T$ leads to the generation of molecules which are only slight variations on molecules seen in the training data. Generation at high $T$ leads to greater diversity but also higher probability of  nonsensical results.

\subsubsection{Optimization with RNNs using reinforcement learning}
Neil et al\ introduced a simple method for repeated MLE which biases generation towards molecules with good properties, which they call \textit{HillClimb-MLE}.\cite{neil2018exploring} Starting with a model that has been trained via MLE on the training data, they generate a large set of SMILES sequences. They then calculate a reward function $R(S)$ for each sequence $S$ generated and find the subset of $N'$ generated molecules with the highest rewards. This subset is used to retrain the model with MLE, and the process is repeated. Each time a new subset of $N'$ generated molecules is determined, it is concatenated on the previous set, so the amount of data being used for MLE grows with each iteration. As this process is repeated the model begins to generate molecules which return higher and higher values from $R(S)$. 

A more common technique is to use \textit{reinforcement learning} after MLE pretraining to fine tune the generator to produce molecules with high reward. The problem of sequence generation can be recast as a reinforcement learning problem with a discrete action space. At each timestep time $t$, the current state of the ``environment'' is the sequence generated so far is $(s_0, \cdots, s_t)$ and the action $a$ is next token to be chosen, $a = s_{t+1}$. The goal of reinforcement learning is to maximize the expected return $G_T$ for all possible start states $s_0$. The return function $G_t = \sum_{i=t}^T R_i $ simply sums the rewards over the length of time the agent is active, which is called an episode. Mathematically the optimization problem reinforcement learning tries to solve is expressed as: 
\begin{equation}\label{RLobjective}
      \max\limits_\theta J(\theta) = \Bbb{E}[G_T|s_0, \theta] 
\end{equation}
where $\theta$ are the parameters of the model. In our case, one episode corresponds to the generation of one molecule, there is only one start state, (the `GO' character) and $R_t = 0$ until the end-of-sequence (`EOS') token is generated or the max string length is reached. If $T$ denotes the max length of the SMILES string then only $R_T$ is non-zero and therefore $G_t = R_T$ for all $t$. The state transition is deterministic (i.e. $p^a_{s,s'} = 1$ for the next state $S_{1:t+1}$ if the current state is $S_{1:t}$ and the action $a = s_{t+1}$, while for other states $s''$, $p^a_{s,s''} = 0$). Because of these simplifications, eqn.\ \ref{RLobjective} assumes a simple form: 
\begin{equation}
    J(\theta) = R_T \sum\limits_{t=0}^{T} \pi_\theta(a_t|s_t)
\end{equation}
Here the policy model $\pi_\theta(a|s)$ gives the probability for choosing the next action given the current state. In our case: 
\begin{equation}
    \pi_\theta(a_t|s_t) = \pi_\theta (s_{t+1} | S_{1:t})
\end{equation} 

There are many reinforcement learning methods, but broadly speaking they can be broken into value learning and policy learning methods. Most work so far has used variants of the REINFORCE algorithm,\cite{Williams1992Simple} a type of policy learning method which falls into the class of policy gradient methods. It can be shown that for a run of length $T$ the gradient of $J(\theta)$ (eqn. \ref{RLobjective}) is: 
\begin{equation}
    \nabla J (\theta) = \mathbb{E}\left[ G_t \frac{\nabla_\theta \pi_\theta(a_t|y_{1:t-1})  }{  \pi_\theta(a_t|y_{1:t-1})} \right]
\end{equation}
Computing the exact expectation of $G_t$ for all possible action sequences is impossible, so instead the $G_t$ from a single run is used before each gradient update. This is sometimes referred to as a ``Monte-Carlo'' type approach. Fortunately, this process can be parallelized by calculating multiple gradient updates on different threads before applying them. Neil et al.\ recently tested several newer reinforcement learning algorithms\textemdash Advantage Actor-Critic (AAC) and Proximal Policy Optimization (PPO), where they report superior performance over REINFORCE (PPO $>$ AAC $>$ REINFORCE). Interestingly, they find \textit{Hillclimb-MLE} is competitive with and occasionally superior to PPO.\cite{neil2018exploring}

Olivecrona et al.\ argue that policy learning methods are a more natural choice for molecular optimization because they can start with a pre-trained generative model, while value-function learning based methods  cannot.\cite{Olivecrona2017} Additionally, most policy learning methods have been proven to lead to an optimal policy and the resulting generative models are faster to sample.\cite{Olivecrona2017} In contrast, Zhou et al.\ argue that value function learning methods are superior in part because policy gradient methods suffer from issues with high variance in the gradient estimates during training.\cite{Zhou2018arXiv}  

Empirically it has been found that using RL after MLE can cause the generated model to ``drift'' too far, causing important information about viable chemical structures learned during MLE to be lost. This can take the form of highly unstable structures being generated or invalid SMILES. One solution is to ``augment'' the reward function with the likelihood:\cite{Olivecrona2017,Jaques2017ICML}
\begin{equation}
    R'(S) = [ \sigma R(S) +  \log P_{\ff{prior}}(S) - \log P_{\ff{current}}(S)]^2
\end{equation}
Other possibilities are explored by Olivecrona et al.\cite{Olivecrona2017} Fundamentally, whether ``drift'' during RL training becomes an issue depends on the details of the reward function\textemdash if the reward function is good, drift should be in a good direction. Recently Zhou et al.\ sought approaches that circumvent MLE when training. In their RL based approach for molecular optimization, they do not use an RNN or any pre-trained generative model and instead use pure RL training.\cite{Zhou2018arXiv} The RL agent works directly on constructing molecular graphs, taking actions such as atom/bond addition and atom removal. The particular approach they use is deep-$Q$ learning, which incorporates several recent innovations that were developed at \textit{DeepMind} and elsewhere.\cite{Mnih2015humanlevel} 
Jaques et al.\ have also explored the application of deep $Q$-learning and $G$-Learning to molecular optimization.\cite{Jaques2017ICML} Reinforcement learning is a rapidly developing field, and there remain many recent advancements such as new attention mechanisms which have not yet been tested in the domain of molecular optimization. 

To give a flavor of what applications have been demonstrated, we will breifly present some representative works using RNNs. Bjerrum \& Threfal explore using an architecture consisting of 256 LSTM cells followed by a ``time-distributed dense'' output layer.\cite{Bjerrum2017arxiv} Their network achieved a SMILES validity rates of $98$\% and the property distributions for the properties tested matched the property distributions found in the training data (some key properties they looked at were synthetic accessibility score, molecular weight, LogP, and total polar surface area). Popova et al.\ have shown how an RNN trained for generation can be further trained with reinforcement learning to generate molecules targeted towards a specific biological function - in their case they focused on the decree to which molecules bind and inhibit the JAK2 enzyme, for which much empirical data exists. They showed how their system could be used to either maximize or minimize inhibition with JAK2 and also independently discovered 793 commercially available compounds found in the ZINC database.\cite{Popova2018} In a similar vein, Segler et al.\ fine tune an RNN to generate a ``focused library'' of molecules which are likely to target the 5-HT$_{\ff{2A}}$ receptor.\cite{Segler2017:120} Finally, Olivecrona et al.\ show how a generative model can be fine tuned to generate analogs of a particular drug (Celecoxib) or molecules which bind to the type 2 dopamine receptor.\cite{Olivecrona2017} 

\subsection{Autoencoders}
\begin{figure*}[h]
  \centering
  \includegraphics[width=19cm]{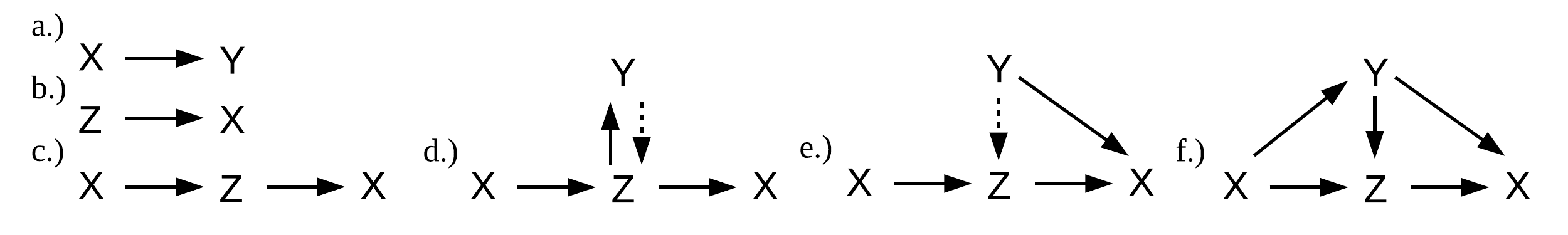}
  \caption{Different deep learning approaches visualized as graphical models. Figure adapted from Kang et al.\cite{Kang2018} Solid lines represent explicit conditional dependencies while dashed lines represent implicit conditional dependencies (arising from the relationship between $X$ and $Y$ inherent in the training data) for which disentanglement may be desired. a.) regression (property prediction) b.) direct generative model c.) autoencoder d.) supervised autoencoder, type 1\cite{GmezBombarelli2018:268} e.) supervised autoencoder, type 2\cite{Polykovskiy2018asap,Lim2018springerASAP} f.) supervised/semisupervised autoencoder, type 3 \cite{Kang2018} }
  \label{graphicalmodels}
\end{figure*}

In 2006 Hinton and Salakhutdinov showed how advances in computing power allowed for the training of a deep autoencoder which was capable of beating other methods for document classification.\cite{Hinton2006:504} The particular type of neural network they used was a stack of restricted Boltzmann machines, an architecture which would later be called a ``deep Boltzmann machine".\cite{Salakhutdinov2009:448} While deep Boltzmann machines are theoretically powerful, they are  computationally expensive to train and impractical for many tasks. In 2013 Kingma et al.\ introduced the variational autoencoder (VAE), \cite{Kingma2013:arxiv} which was used in 2016 by Bombarelli et al.\  to create the first machine learning based generative model for molecules.\cite{GmezBombarelli2018:268} 


\subsubsection{Variational autoencoders (VAEs)}
VAEs are derived mathematically from the theory of variational inference and are only called autoencoders because the resulting architecture has the same high level structure as a classical autoencoder. VAEs are fundamentally a latent variable model $p(\bs{x}, \bs{z}) = p_\theta(\bs{x}|\bs{z})p(\bs{z})$ which consists of latent variables $\bs{z}$ drawn from a pre-specified prior $p(\bs{z})$ and passed into a decoder $p_\theta(\bs{x}|\bs{z})$ parametrized by parameters $\theta$. To apply maximum likelihood learning to such a model we like to maximize the probability of each observed datapoint $p(\bs{x}) = \int p_\theta(\bs{x}|\bs{z})p(\bs{z}) d\bs{z}$ for all datapoints in our training data. However for complicated models with many parameters $\theta$ (like neural networks) this integral is intractable to compute. The method of variational inference instead maximizes a lower bound on $\log p(\bs{x})$: \begin{equation}\label{lowerbound}
    \log p(\bs{x}) \geq \mathbb{E}_{\bs{z}\sim q_\phi (\bs{z}|\bs{x})}\left[ \log \frac{ p_\theta(\bs{x}|\bs{z})p(\bs{z})}{q_\phi (\bs{z}|\bs{x})} \right]
\end{equation}
where $q_\phi (\bs{z}|\bs{x})$ is an estimate of posterior distribution $p(\bs{z}|\bs{x}) = p_\theta(\bs{x}|\bs{z})p(\bs{z})/p(\bs{x})$. The right hand side of eqn.~\ref{lowerbound} is called the ``negative variational free energy'' or ``evidence lower bound'' (ELBO) and can be written as:  
\begin{equation}\label{ELBO}
   \mathcal{L}_{\theta, \phi}(\bs{x})  = 
   \mathbb{E}_{\bs{z}\sim q_\phi (\bs{z}|\bs{x})}[ \log p_\theta(\bs{x})] -  D_{\ff{KL}}(q_\phi(\bs{z}|\bs{x}),p_\theta(\bs{z}|\bs{x}))
\end{equation}
Here we encounter the Kullback-Leibler (KL) divergence: 
\begin{equation}\label{KLdiv}
    D_{\ff{KL}}(q(\bs{z}),p(\bs{z})) \equiv \int  q(\bs{z}) \log \frac{q(\bs{z})}{p(\bs{z})}  d\bs{z}
\end{equation}
After several manipulations, eqn.\ \ref{ELBO} can be written as 
\begin{equation}\label{ELBO2}
    \begin{aligned} 
        \mathcal{L}_{\theta, \phi}(\bs{x})  &= \mathbb{E}_{\bs{z}\sim q_\phi (\bs{z}|\bs{x})}[ \log p(\bs{z},\bs{x})] + H[q_\phi (\bs{z}|\bs{x})] \\
                 &=  \mathbb{E}_{\bs{z}\sim q_\phi (\bs{z}|\bs{x})}[ \log p_\theta(\bs{x}|\bs{z})] - D_{\ff{KL}}[q_\phi (\bs{z}|\bs{x}), p(\bs{z}) ]
     \end{aligned} 
\end{equation}
where $H$ is the (differentiable) entropy. The loss function for the variational autoencoder for examples $\bs{x}$ in our training dataset $\mathcal{Z}$ is: \begin{equation}\label{VAEloss}
    L_{\theta,\phi} = \sum\limits_{\bs{x}\in\mathcal{Z}} - \mathcal{L}_{\theta, \phi}(\bs{x}) 
\end{equation}

In a VAE, during training first $q_\phi (\bs{z}|\bs{x})$ (the encoder) generates a $\bs{z}$. Then the decoder $p_\theta(\bs{x}|\bs{z})$ model attempts to recover $\bs{x}$. Training is done using backpropagation and the loss function (eqn.\ \ref{VAEloss}) which tries to maximize $\mathcal{L}(\bs{x},\theta, \phi) $. This  corresponds to maximizing the chance of reconstruction $p_\theta(\bs{x}|\bs{z})$ (the first term) but also minimizing the KL-divergence between $q_\phi(\bs{z}|\bs{x})$ and the prior distribution $ p(\bs{z})$. Typically the prior is chosen to be a set of independent unit normal distributions and the encoder is assumed to be a factorized multidimensional normal distribution: 
\begin{equation}
    \begin{aligned}
     p(\bs{z})  &= \mathcal{N}(\bs{z}, 0, \bs{I}) \\
     q_\phi (\bs{z}|\bs{x})& = \mathcal{N}(\bs{z}, \bs{\mu}(\bs{x}), \mbox{diag}(\bs{\sigma}^2(\bs{x}))
    \end{aligned} 
\end{equation}
The encoder $q_\phi (\bs{z}|\bs{x})$ is typically a neural network with parameters $\phi$ used to find the mean and variance functions in $\mathcal{N}(\bs{z}, \bs{\mu}(\bs{x}),  \mbox{diag}(\bs{\sigma}^2(\bs{x}))$. The resulting ``Gaussian VAE'' has the advantage that the KL-divergence can be computed analytically. The parameters $\theta$ and $\phi$ in the decoder and encoder are all learned via backpropagation. There are several important innovations which have been developed to streamline backpropagation and training which are described in detail elsewhere.\cite{Kingma2013:arxiv, GoodfellowBook2016, Mehta2018arxiv}  

There are several reasons that variational autoencoders perform better than classical autoencoders. Since the latent distribution is probabilistic, this introduces noise which intuitively can be seen as a type of regularization that forces the VAE to learn more robust representations. Additionally, specifying that the latent space must be a Gaussian leads to a much smoother latent space which makes optimization much easier and also leads to fewer ``holes'' in the distribution corresponding to invalid or bad molecules. VAEs therefore are useful for interpolation between points corresponding to molecules in the training data.

In the molecular autoencoder of G\'{o}mez-Bombarelli et al.\ each SMILES $\bs{x}$ is converted to a one-hot representation and a convolutional neural network is used to find the parameters of the Gaussian distribution $q_\phi(\bs{z}|\bs{x})$.\cite{GmezBombarelli2018:268} 
The decoder in the molecular autoencoder is an RNN, but in contrast to pure RNN models, where high rates of valid SMILES generation have been reported (94-98~\%),\cite{neil2018exploring,Segler2017:120,Bjerrum2017arxiv} the molecular autoencoder generates far fewer valid SMILES. The valid SMILES rate was found to vary greatly between $\approx$ 75\% for points near known molecules to only 4\% for randomly selected latent points.\cite{GmezBombarelli2018:268} Kusner et al.\ report an average valid decoding rate of only 0.7\% using a similar VAE architecture.\cite{KusnerArxiv2017} These low decoding rates are not a fatal issue however simply because a validity checker (such as found in RDKit) can easily be used to throw out invalid SMILES during generation. However, the low rate of validity suggests fundamental issues in quality of the learned latent representation. As mentioned previously, higher rates of SMILES validity have been achieved by representing SMILES in terms of rules from a context-free grammar (CFG).\cite{KusnerArxiv2017, Dai2018arxivSDVAE} Kusner et al.\ achieved somewhat higher rates of SMILES generation using a CFG (7.2\%, as described in the supplementary information of Kusner et al.\cite{KusnerArxiv2017}). Further work by Dai et al.\ added additional ``semantic'' constraints on top of a CFG yielding a higher rate of valid SMILES (43.5\%).\cite{Dai2018arxivSDVAE} Janz et al.\ recently proposed using Bayesian active learning as a method of forcing models to learn what makes a sequence valid, and incorporating this into RNNs in VAEs could lead to higher valid decoding rates.\cite{Janz2017arxiv}

\subsubsection{Adversarial autoencoders (AAEs)}
Adversarial autoencoders are similar to variational autoencoders, but differ in the means by which they regularize the latent distribution by forcing it to conform to the prior $p(\bs{z})$.\cite{Makhzani2016AAE} Instead of minimizing KL-divergence metric to enforce the generator to output a latent distribution corresponding to a prespecified prior (usually a normal distribution), they use adversarial training with a discriminator $D$ whose job is to distinguish the generator's latent distribution from the prior. The discriminator outputs a probability $ p \in (0,1)$ which predicts the probability samples it sees are from the prior. The objective of the discriminator is \textit{maximize} the following: 
\begin{equation}
   \mathcal{L}_{\ff{adv}}=  \mathbb{E}_{\bs{x}\sim p_d}[ \log D( (q_\Theta(\bs{z}|\bs{x}))] + \mathbb{E}_{\bs{x}\sim p_{\bs{z}}}[ \log ( 1 - D( \bs{z}))] 
\end{equation}
The overall objective function for the AAE to \textit{minimize} can be expressed as
\begin{equation}\label{AAEloss}
         L_{\theta,\phi} = \sum\limits_{\bs{x}\in\mathcal{Z}} -\mathbb{E}_{\bs{x}\sim p_d  }[ \log p_\theta(\bs{x}|q_\phi (\bs{z}|\bs{x}))] - \mathcal{L}_{\ff{adv}} 
\end{equation}

\subsubsection{Supervised VAEs/AAEs for property prediction \& optimization}

In supervised VAEs, target properties $\bs{y}$ for each molecule are incorporated into the generator in addition to the SMILES strings or other molecular representation. Figure \ref{graphicalmodels} shows several different ways this can be done, representing the generative models as graphical models. Everything we discuss in this section can also be applied to AAEs,\cite{Makhzani2016AAE} but we restrict our discussion to VAEs for simplicity. 

In the work by G\'{o}mez-Bombarelli et al.\ they attached a neural network (multilayer perceptron) to the latent layer and jointly trained the neural network to predict property values $y$ and the VAE to minimize reconstruction loss. One unique property they optimize after training such a system is the predicted HOMO-LUMO gap, which is important for determining a molecule's utility in organic solar cells. The advantage of supervised VAEs is that the generator learns a good latent representation both for property prediction and reconstruction. With the property predictor trained, it becomes possible to do property optimization in the latent space, by either using Gaussian process optimization or gradient ascent. Interestingly, in supervised VAEs a particular direction in the latent space always became correlated with the property value $y$, while this was never observed in unsupervised VAEs. When one desires to do optimization, G\'{o}mez-Bombarelli et al.\ argue for using a Gaussian process model as the property predictor instead of a neural network, because it generates a smoother landscape.\cite{GmezBombarelli2018:268} The specific procedure they used was to first train a VAE using supervised training with a neural network property predictor and then train a Gaussian process model separately using the latent space representations of the training molecules as input. They then use the Gaussian process model for optimization, and they showed it was superior to a genetic optimization algorithm and random search in the latent space. Since that work, several other researchers have used Gaussian process regression to perform optimization in the latent space of a VAE.\cite{Ikebata2017:379,Griffiths2017arxiv,Samanta2018:arxiv}

Two other types of supervised VAEs are shown in figure \ref{graphicalmodels}, which we call ``type 2'' and ``type 3''. Unlike the autoencoder frame work discussed in the previous section, these two types of autoencoder can be used for conditional generation. In ``type 3'' supervised VAEs the ELBO term in the objective function (eqn. \ref{ELBO2}) becomes:\cite{Kang2018}
\begin{equation}
       \mathbb{E}_{\bs{z}\sim q_\phi (\bs{z}|\bs{x},\bs{y})}[ \log p_\theta(\bs{x}|\bs{y},\bs{z})] - D_{\ff{KL}}[q_\phi (\bs{z}|\bs{x},\bs{y}) || p(\bs{z},\bs{y}) ]
\end{equation}
Kang et al.\ assume that the property values have a Gaussian distribution. 
Type 3 VAEs are particularly useful when  $\bs{y}$ is known for only some of the training data (a semi-supervised setting). In semi-supervised VAEs, the generator is tasked with predict $\bs{y}$ and is trained on the molecules where $\bs{y}$ is known and makes a best guess prediction for the rest. In effect, when  $\bs{y}$ is not known, it becomes just another latent variable and a different objective function is used (for details, see Kang et al.\cite{Kang2018}). 

In Type 2 VAEs, property data  $\bs{y}$ is embedded directly into the latent space during training.\cite{Lim2018springerASAP,Polykovskiy2018asap} Supervised and semi-supervised VAEs can both be used for conditional sampling, and thus are sometimes called ``conditional VAEs''. In the traditional way of doing conditional sampling, $\bs{y}$ is specified and then one samples from the prior $p(\bs{z})$. Then one samples from the generator $p_\theta(\bs{x}|\bs{y},\bs{z})$. In the case of Type 1 and Type 2 VAEs, however, there is an issue pointed out by Polykovskiy et al.\, which they call ``entanglement''.\cite{Polykovskiy2018asap} The issue is that when sampling we assumed that $p(\bs{z})$ is independent of $p(\bs{y})$. However, the two distributions are actually ``entangled'' by the implicit relationship between $x$ and  $\bs{y}$ which is in the training data (this is indicated by a dashed line in figure \ref{graphicalmodels}). For consistency, one should be sampling from $p(\bs{z}|\bs{y})$. Polykovskiy et al.\ developed two ``disentanglement'' approaches to ameliorate this issue: learning $p(\bs{z}|\bs{y})$ and forcing all $p(\bs{z}|\bs{y})$ to match $p(\bs{z})$.\cite{Polykovskiy2018asap} 

When generating molecules with an RNN, we previously discussed sampling from the model's distribution by simply running the model and taking either the token with the maximum probability or using a multinomial sampler at each step of the sequence generation. When sampling from the generator of a conditional VAE, we wish to know what the model says is the likely molecule given $\bs{y}$ and $\bs{z}$, since we are interested in focusing on the molecules the model predicts are most likely to be associated with a particular set of properties:  
\begin{equation}\label{argmaxx}
    \hat{\bs{x}} = \argmax_{\bs{x}} p_\theta(\bs{x}|\bs{y},\bs{z})
\end{equation}
Taking the most likely token at each step (the ``greedy'' method) is only a rough approximation to $\hat{\bs{x}}$. Unfortunately, completing the optimization in eqn.\ \ref{argmaxx} is a computationally intractable problem because the space of sequences grows exponentially with the length of the sequence. However, a variation on the greedy method called ``beam search'' can be used to get an approximation of $\hat{\bs{x}}$.\cite{Sutskever2014arXiv,Kang2018} In brief, beam search keeps the top $K$ most likely (sub)sequences at each step of the generation.

\subsection{Generative adversarial networks (GANs)}
The key idea underlying GANs is to introduce a discriminator network whose job is to distinguish whether the molecule it is looking at was generated by the generative model or came from the training data. In GAN training, the objective of the generative model becomes to try to fool the discriminator rather than maximizing likelihood. There are theoretical arguments and growing empirical evidence showing that GAN models can overcome some of the well known weaknesses of maximum likelihood based training. However, there are also many technical difficulties which plague GAN training and getting GANs to work well typically requires careful hyperparameter tuning and implementation of several non-obvious ``tricks''.\cite{Lucic2018arXivGANsCreatedEqual} GANs are a rapidly evolving research area, and given space limitations we can only touch on several of the key developments here. 

The original paper on GANs (Goodfellow et al\ 2014) introduced the following objective function:\cite{Goodfellow2014:2672}
\begin{equation}\label{eqn:objfun}
    \begin{aligned}
    \min_G \max_D V(D, G) = & \mathbb{E}_{\bs{x}\in p_d(\bs{x})}[\log D(\bs{x})]\\
        & + \mathbb{E}_{\bs{z}\in p_{\bs{z}}(\bs{z})} [\log(1 - D(G(\bs{z}))] 
    \end{aligned}
\end{equation}
Here $p_d(\bs{x})$ is the data distribution. This form of the objective function has a nice interpretation as a two person minimax game. However, this objective function is rarely used in practice for a few reasons. Firstly, as noted in the original paper, this objective function does not provide a very strong gradient signal when training starts because then $\log(1 - D(G(\bs{z}))$ saturates (goes to negative infinity) and the numerical derivative becomes impossible to calculate. Still, understanding this objective function can help understand how generative modeling with GAN training can be superior to maximum likelihood based generative modeling. For a fixed $G$, the optimal discriminator is: 
\begin{equation}
    D^*_G(\bs{x}) = \frac{ p_d(\bs{x}) }{ p_d(\bs{x}) + p_{\theta_{G}}(\bs{x}) }
\end{equation}
If we assume $D=D_G^*$, then the objective function $\mathcal{C}(G)$ for the generator can be expressed as:\cite{Goodfellow2014:2672}
\begin{equation}\label{GANobjective}
    \mathcal{C}(G) = -\log (4) + 2 D_{\ff{JS}}(p_d, p_{\theta_{G}})
\end{equation}
Where $D_{\ff{JS}}(p_d, p_{\theta_{G}})$ is the Jensen-Shannon divergence:
\begin{equation}\label{JSdivergence} 
    \begin{aligned} 
    D_{\ff{JS}}(p_d, p_{\theta_{G}}) = & \frac{1}{2}  D_{\ff{KL}}\left(p_d\bigg{|}\bigg{|} \frac{p_d+p_{\theta_{G}}}{2}\right)  \\
    & + \frac{1}{2}  D_{\ff{KL}}\left(p_{\theta_{G}}\bigg|\bigg| \frac{p_d+p_{\theta_{G}}}{2}\right) 
    \end{aligned} 
\end{equation}
Here $D_{\ff{KL}}(p,q)$ is the Kullback-Leibler (KL) divergence. Maximizing the log-likelihood is equivalent to minimizing the forward KL divergence $ D_{\ff{KL}}(p_d ,p_{\theta_{G}})$.\cite{Mehta2018arxiv} To better understand what this means, we can rewrite the equation for KL divergence (eqn.\ \ref{KLdiv}) in a slightly different way:
\begin{equation}\label{KLdiv2}
    D_{\ff{KL}}(p_d ,p_{\theta_G} ) = \int  p_d (\bs{z}) ( \log p_d(\bs{z}) - \log p_{\theta_{G}}(\bs{z}) ) d\bs{z}
\end{equation} 
This shows us that KL divergence captures the difference between $p_d$ and $p_{\theta_G}$ weighted by $p_d$. Thus one of the weaknesses of maximum likelihood is that $p_{\theta_G}$ may have significant deviations from $p_d$ when $p_d \approx 0$. To summarize, the forward KL divergence ($ D_{\ff{KL}}(p_d ,p_{\theta_{G}})$) punishes models that underestimate the data distribution, while the reverse KL divergence ($ D_{\ff{KL}}(p_{\theta_{G}}, p_d)$) punishes models that overestimate the data distribution. Therefore we see that eqn. \ref{GANobjective}, which contains both forward and backwards KL terms, takes a more ``balanced'' approach than maximum likelihood, which only optimizes forward KL divergence. Optimizing reverse KL divergence directly requires knowing an explicit distribution for $p_d$, which usually is not available. In effect, the discriminator component of the GAN works to learn $p_d$, and thus GANs provide a way of overcoming this issue. 

As noted before, the GAN objective function given in eqn.\ \ref{eqn:objfun}, however, does not provide a good gradient signal when training first starts since typically $p_d$ and $p_{\theta_{G}}$ have little overlap to begin with. Empirically, this occurs because data distributions typically lie on a low dimensional manifold in a high dimensional space, and the location of this manifold is not known in advance. The Wasserstein GAN (WGAN) is widely accepted to provide a better metric for measuring the distance between $p_d$ and $p_{\theta_{G}}$ than the original GAN objective function and results in  faster and more stable training.\cite{Arjovsky2017arxiv} The WGAN is based on the Wasserstein metric (also called the ``Earth mover's distance") which can be informally understood by imagining the two probability distributions $p_d$ and $p_{\theta_{G}}$ to be piles of dirt, and the distance between them to be the number of buckets of dirt that need to be moved to transform one to the other, times the sum of the distances each bucket must be moved. Mathematically this is expressed as: 
\begin{equation}
    W(p, q) = \inf_{\gamma \in \Pi(p, q)} \Bbb{E}_{(x,y) \in \gamma} || x - y ||
\end{equation}
$\Pi(x, y)$ can be understood to be the optimal ``transport plan'' explaining how much probability mass is moved from $x$ to $y$. 
Another feature of the WGAN is the introduction of a ``Lipschitz constraint'' which clamps the weights of the discriminator to lie in a fixed interval. The Lipschitz constraint has been found to result in a more reliable gradient signal for the generator and improve training stability. Many other types of GAN objective function have been developed which we do not have room to discuss here. For a review of the major GAN objective functions and latest techniques, see Kurach et al.\cite{Kurach2018arXivGANreview} 

Several papers have emerged so far applying GANs to molecular generation\textemdash Guimares et al.\ (ORGAN),\cite{Guimaraes2017ORGANarxiv} S\'{a}nchez-Lengling et al.\ (ORGANIC), \cite{Sanchez-Lengeling2017ORGANICchemrxiv} De Cao \& Kipf (MolGAN),\cite{deCao2018:arxiv,deCao2018molganICML} and Putin et al.\ (RANC, ATNC).\cite{Putin2018:1194,Putin2018asap} The Objective-Reinforced GAN (ORGAN) of Guimares et al.\ uses the SMILES molecular representation and an RNN (LSTM) generator and a CNN discriminator.\cite{Guimaraes2017ORGANarxiv} The architecture of the ORGAN is taken from the SeqGAN of Yu et al.\ \cite{Yu2017:2852} and uses a WGAN. In ORGAN, the GAN objective function is modified by adding an additional ``objective reinforcement'' term to the generator RNN's reward function which biases the RNN to produce molecules with a certain objective property or set of objective properties. Typically the objective function returns a value $R(S) \in [0, 1]$. The reward for a SMILES string $S$ becomes a mixture of two terms: 
\begin{equation}
    R(S) = \lambda D(S) + (1-\lambda)) R(S)
\end{equation}
where $\lambda \in [0, 1]$ is a tunable hyperparameter which sets the mixing between rewards for fooling the discriminator and maximizing the objective function. The proof of concept of the ORGAN was demonstrated by optimizing three quick-to-evaluate metrics which can be computed with RDKit\textemdash druglikeliness, synthesizability, and solubility. Proof of concept applications of the ORGAN have been demonstrated in two domains - firstly for drug design it as shown how ORGAN can be used to mazimize Lapinksi's rule of five metric as well as the quantitative estimate of drug likeliness metric. The second application for which ORGAN was demonstrated is maximizing the power conversion efficiency (PCE) of molecules for use in organic photovoltaics, where PCE is estimated using a machine learning based predictor that was previously developed.\cite{Sanchez-Lengeling2017ORGANICchemrxiv}


\subsubsection{The perfect discriminator problem and training instabilities}
GAN optimization is a saddle point optimization problem, and such problems are known to be inherently unstable. If gradients for one part of the optimization dominate, optimizers can run or ``spiral'' away from the saddle point so that either the generator or the discriminator achieves a perfect score. The traditional approach to avoiding the perfect discriminator problem, taken by Guimares et al.\ and others, is to do additional MLE pretraining with the generator to strengthen it and then do $m$ gradient updates for the generator for every one gradient update for the discriminator. In this method, $m$ must be tuned to balance the discriminator and generator training. A different, more dynamic method for balancing the discriminator and generator was invented by Kardurin et al.\ in their work on the DruGAN AAE.\cite{Kadurin2017:3098} They introduce a hyperparameter $0.5 < p < 1$ which sets the desired ``discriminator power''. Then, after each training step, if the discriminator correctly labels samples from the generator with probability less than $p$, the discriminator is trained, otherwise the generator is trained. Clearly $p$ should be larger than 0.5 since the discriminator should do better than random chance in order to challenge the generator to improve. Empirically, they found $p = 3/5$ to be a good value. 

Putin et al.\ show that the ORGANIC model\cite{Sanchez-Lengeling2017ORGANICchemrxiv} with its default hyperparameters suffers from a perfect discriminator problem during training, leading to a plateauing of the generator's loss.\cite{Putin2018:1194} To help solve these issues, Putin et al.\ implemented a differentiable neural computer (DNC) as the generator.\cite{Putin2018:1194} The DNC (Graves et al, 2016)\cite{Graves2016:471} is an extension of the neural Turing machine (Graves et al\, 2014)\cite{Graves2014NTMarxiv} that contains a differentiable memory cell. A memory cell allows the generator to memorize key SMILES motifs, which results in a much ``stronger'' generator. They found that the discriminator never achieves a perfect score when training against a DNC. The strength of the DNC is also shown by the fact that it has a higher rate of valid SMILES generation vs.\ the ORGAN RNN generator (76\% vs.\\ 24\%) and generates SMILES that are on average twice as long as the SMILES generated by ORGAN. 
In a subsequent work, Putin et al.\ also introduced the adversarial threshold neural computer, another architecture with a DNC generator.\cite{Putin2018asap}

Another issue with GANs is mode collapse, where the generator only generates a narrow range of samples. In the context of molecules, an example might be a generator that only generates molecules with carbon rings and less than 20 atoms. 

\section{Metrics and reward functions}
A key issue in deep generative modeling research is how to quantitatively compare the performance of different generative models. More generally a decline in rigor in the field of deep learning as a whole has been noted by Sculley, Rahimi and others.\cite{Sculley2018ICML} While the recent growth in the number of researchers in the field has obvious benefits, the increased competition that can result from such rapid growth disincentivizes taking time for careful tuning and rigorous evaluation of new methods with previous ones. Published comparisons are often subtly biased towards demonstrating superior performance for technically novel methods vs.\ older more conventional methods. A study by Lucic et al.\ for instance found that in the field of generative adversarial networks better hyperparameter tuning and training lead to most recently proposed methods reaching very similar results.\cite{Lucic2018arXivGANsCreatedEqual,Henderson2018DeepRL} Similarly, Melis et al.\ found that with proper hyperparameter tuning a conventional LSTM architecture could beat several more recently proposed methods for natural language modeling.\cite{Melis2017arXiv} At the same time, there is a reproducibility crisis afflicting deep learning\textemdash codes published side-by-side with papers often give different results than what was published,\cite{Kurach2018arXivGANreview} and in the field of reinforcement learning it has been found that codes which purport to do the same thing will give different results.\cite{Henderson2018DeepRL} The fields of deep learning and deep generative modeling are still young however, and much work is currently underway on developing new standards and techniques for rigorously comparing different methods. In this section we will discuss several of the recently proposed metrics for comparing generative models and the closely related topic of reward function design for molecular optimization.

\subsection{Metrics for scoring generative models}
Theis et al.\ discuss three separate approaches\textemdash log-likelihood, estimating the divergence metric between the training data distribution $p(x)$ and the model's distribution $q(x)$, and human rating by visual inspection (also called the ``visual Turing test'') .\cite{Salimans2016InceptionScore,Theis2016ICLR} Interestingly, they show that these three methodologies measure different things, so good performance under one does not imply good performance under another.\cite{Theis2016ICLR} 

The ``inception score'' (IS) uses a third-party neural network which has been trained in a supervised manner to do classification.\cite{Salimans2016InceptionScore} In the original IS, Google's Inception network trained on ImageNet was used as the third-party network. IS computes the divergence between the distribution of classes predicted by the third-party neural network on generated molecules with the distribution of classes predicted for the dataset used to train the neural network. The main weakness of IS is that much information about the quality of images is lost by focusing only on classification labels. The Fr\'{e}chet Inception Distance (FID) builds off the IS by comparing latent vectors obtained from a late-stage layer of a third-party classification network instead of the predictions.\cite{Heusel2017FrechetID} Inspired by this, Preuer et al.\ created the Fr\'{e}chet ChemNet Distance metric for evaluating models that generate molecules.\cite{Preuer2018Frechet} Unfortunately, there is a lack of agreement on how to calculate the FID\textemdash some report the score by comparing training data with generated data, while others report the score comparing a hold out test set with the generated data.\cite{Kurach2018arXivGANreview} Comparing with test data gives a more useful metric which measures generalization ability, and is advocated as a standard by Kurach et al.\cite{Kurach2018arXivGANreview} 

In the world of machine learning for molecular property prediction, \textit{MoleculeNet} provides a benchmark to compare the utility of different regression modeling techniques across a wide range of property prediction problems.\cite{Wu2018MoleculeNet} Inspired by \textit{MoleculeNet}, Polykovskiy and collaborators have introduced the MOlecular SEtS (MOSES) package to make it easier to build and test generative models.\cite{Polykovskiy2018arXivMOSES} To compare the output of generative models, they provide functions to compute Fr\'{e}chet ChemNet Distance, internal diversity, as well as several metrics which are of general importance for pharmaceuticals: molecular weight, logP, synthetic accessibility score, and the quantitative estimation of drug-likeness. In a similar vein, Benhenda et al.\ have released the \textit{DiversityNet} benchmark, which was also (as the name suggests) inspired by \textit{MoleculeNet}.\cite{diversitynet2018} Finally, another Python software package called \textit{GuacaMol} has also been released which contains 5 general purpose benchmarking methods and 20 ``optimization specific'' benchmarking methods for drug discovery.\cite{Brown2018GuacaMol} One unique feature of \textit{GuacaMol} is the ability to compute KL-divergences between the distributions from generated molecules and training molecules for a variety of physio chemical descriptors. 

Recently in the context of generative modeling of images with GANs, Im et al.\ have shown significant pitfalls to using the Inception Distance metric.\cite{jiwoong2018ICLRquantitatively} As an alternative, they suggest using the same type of divergence metrics that are used during GAN training. This method has been explored recently to quantify generalization performance of GANS\cite{Gulrajani2018towardsICLR} and could be of use to the molecular modeling community.  

\subsection{Reward function design}
A good reward function is often important for molecular generation and essential for molecular optimization. The pioneering molecular autoencoder work resulted in molecules which were difficult to synthesize or contained highly labile (reactive or unstable) groups such as enamines, hemiaminals, and enol ethers which would rapidly break apart in the body and thus were not viable drugs.\cite{LoweBlogPost} Since then, the development of better reward functions has greatly helped to mitigate such issues, but low diversity and novelty remains an issue.\cite{Benhenda2017arxiv,Yoshikawa2018arxiv,Panteleev2018:Recent} After reviewing the work that has been done so far on reward function design, we conclude that good reward functions should lead to generated molecules which meet the following desiderata:
\begin{enumerate}
    \item \textbf{Diversity}\textemdash the set of molecules generated is diverse enough to be interesting. 
    \item \textbf{Novelty}\textemdash the set of molecules does not simply reproduce molecules in the training set.
    \item \textbf{Stability}\textemdash the molecules are stable in the target environment and not highly reactive.
    \item \textbf{Synthesizability}\textemdash the molecules can actually be synthesized. 
    \item \textbf{Non-triviality}\textemdash the molecules are not degenerate or trivial solutions to maximizing the reward function. 
    \item \textbf{Good properties}\textemdash the molecules have the properties desired for the application at hand. 
\end{enumerate}  

\subsubsection{Diversity \& novelty} 
A diversity metric is a key component of any reward function, especially when using a GAN, where it helps counter the issue of mode collapse to a non-diverse subset. Given a molecular similarity metric between two molecules $T(x_1, x_2) \in [0,1]$ the diversity of a generated set $\mathcal{G}$ can be defined quite simply as: 
\begin{equation}\label{diversity} 
    r_{\ff{diversity}} = 1 - \frac{1}{|\mathcal{G}|} \sum\limits_{(x_1, x_2) \in \mathcal{G}\times\mathcal{G}} D(x_1, x_2)
\end{equation}
A popular metric is the Tanimoto similarity between two extended-connectivity fingerprint bit vectors.\cite{Polykovskiy2018arXivMOSES} Since the diversity of a single molecule does not make sense, diversity rewards are calculated on mini-batches during mini-batch stochastic gradient descent training. Eqn.\ \ref{diversity} is called ``internal diversity''. An alternative which compares the diversity of the generated set with the diversity of the training data is the nearest neighbor similarity (SNN) metric:\cite{Polykovskiy2018arXivMOSES}
\begin{equation}\label{SNN}
    r_{\ff{SSN}} = \frac{1}{|\mathcal{G}|} \sum\limits_{x_G \in \mathcal{G}} \max\limits_{x_D \in \mathcal{D}} D(x_G, x_G)
\end{equation}
Of course, too much diversity can also be an issue. One option is to use the following negative reward which biases the generator towards matching the diversity of the training data:
\begin{equation}
    R_{\ff{diversity mismatch}} = - \left| r^{\ff{generated}}_{\ff{diversity}} - r^{\ff{training}}_{\ff{diversity}}\right|
\end{equation}

Another diversity measure that has been employed is uniqueness, which aims to reduce the number of repeat molecules. The uniqueness reward $ R_{\ff{uniqueness}} \in (0,1]$ is defined as:
\begin{equation}
    R_{\ff{uniqueness}} = \frac{|\mbox{set}(\mathcal{G})|}{|\mathcal{G}|}
\end{equation}
Where $|\mbox{set}(\mathcal{G})|$ is the number of unique molecules in the generated batch  $\mathcal{G}$ and $|\mathcal{G}|$ is the total number of molecules.

Novelty is just as important as diversity since a generator which just generates the training dataset over and over is of no practical utility. Guimares et al.\ define the ``soft novelty'' for a single molecule as:\cite{Guimaraes2017ORGANarxiv} 
\begin{equation}
R_{\ff{novelty}}  =
    \begin{cases}
      1  & \mbox{If $x$ is not in the training set} \\
     0.3 & \mbox{If $x$ is in the training set} \\ 
    \end{cases} 
\end{equation}

When measuring the novelty of molecules generated post-training to get an overall novelty measure for the model, it is important to do so on a hold-out test set $\mathcal{T}$. Then one can look at how many molecules in a set of generated molecules $\mathcal{G}$ appear in $\mathcal{T}$ and use a novelty reward such as:\cite{Segler2017:120}
\begin{equation}
     r_{\ff{novel}}=  1 -  \frac{|\mathcal{G} \cap \mathcal{T}|}{|\mathcal{T}|} 
\end{equation}

which gives the fraction of generated molecules not appearing in the test set. The diversity of the generated molecules and how they compare to the diversity of the training set can also be visualized by generating fingerprint vectors (which typically have dimensionalities of $d>100$) and then projecting them into two dimensions using dimensionality reduction techniques. The resulting 2D point cloud can then be compared with the corresponding points from the training set and/or a hold out test set. There are many possible dimensionality reduction techniques to choose from\textemdash Yoshikawa et al.\cite{Yoshikawa2018arxiv} use the ISOMAP method,\cite{Tenenbaum2000:2319} Merk et al.\cite{Merk2018:1700153} use multidimensional scaling, and Selger et al.\cite{Segler2017:120} use t-SNE projection.\cite{vanDerMaaten2008:2579}

Interpolation between training molecules may be a useful way to generate molecules post-training which are novel, but not too novel as to be unstable or outside the intended property manifold. In the domain of image generation, GANs seem to excel at interpolation vs.\ VAEs, for reasons that are not yet fully understood. For instance with GANs trained on natural images, interpolation can be done between a $\bs{z}$ point corresponding to a frowning man to a point $\bs{z}'$ corresponding to a smiling woman, and all of the intervening points result in images which make sense.\cite{radford2016unsupervisedICLR} Empirically most real world high dimensional datasets are found to lie on a low density manifold.\cite{Domingos2012:78} Ideally, the dimensionality of the latent space $p(\bs{z})$ used in a GAN, VAE, or AAE will correspond to the dimensionality of this manifold. If the dimensionality of $p(\bs{z})$ is higher than the intrinsic dimensionality of the data manifold, then interpolation can end up going ``off manifold'' into so-called ``dead zones''. 
For high dimensional latent spaces with a Gaussian prior, most points will lie on a thin spherical shell. In such cases it has been found empirically that better results can be found by doing spherical linear interpolation or {\it slerp}.\cite{Whitearxiv2016} The equation for \textit{slerp} interpolation between two vectors $\bs{v}_1$ and $\bs{v}_2$ is 
\begin{equation}
    \mbox{\textit{slerp}}(\bs{v}_1, \bs{v}_2, t)  = \frac{ \sin((1 - t)\theta)}{\sin(\theta)} \bs{q}_1 + \frac{\sin(t \theta)}{\sin(\theta)} \bs{q}_2
\end{equation}
where $\theta = \arccos(\bs{v}_1\cdot\bs{v}_2)$ and $0 \leq t \leq 1$ is the interpolation parameter. 

Another option for generating molecules close to training molecules but not too close is to have a reward for being similar to the training data but not too similar. Olivecrona et al.\ use a reward function $R_s(S) \in [-1,1]$ of the form:\cite{Olivecrona2017}
\begin{equation}
    R_s(S) = 1 - 2 \frac{\min (\mbox{Sim}(S,T), k)}{k}
\end{equation}
here $S$ is the input SMILES and $T$ is the target SMILES, and $\mbox{Sim}(S,T) \in [0, 1]$ is similarity scoring function which computes fingerprint-based similarity between the two molecules. $k$ is a tunable cutoff parameter which sets the maximum similarity accepted. This type of reward can be particularly useful for generating focused libraries of molecules very similar to a single target molecule or a small set of ``actives'' which are known to bind to a receptor. Note that most generative models can be run so as to generate molecules close to a given molecule. For instance, with RNNs, one can do ``fragment growing'', which allows molecular designers to explore molecules which share a predefined  scaffold.\cite{Gupta2017:1700111} Similarly, with reinforcement learning one can simply start the agent with a particular molecule and let it add or remove bonds. Finally, with a VAE one can find the latent representation for a given molecule and then inject a small amount of Gaussian noise to generate ``nearby'' molecules.\cite{Harel2018prototype}

\subsubsection{Stability and synthesizability} 
Enforcement of synthesizability has thus far mainly been done using the synthetic accessibility (SA) score developed by Ertl \& Schuffenhauer, \cite{Ertl2009:8} although other synthesizability scoring functions exist.\cite{Podolyan2010:979,Fukunishi2014:3259} The model underlying the SA score was designed and fit specifically to match scores from synthetic chemists on a set of drug molecules, and therefore may be of limited applicability to other types of molecules. When using SA score as a reward in their molecular autoencoder, G\'{o}mez-Bombarelli et al.\ found that it still produced a large number of molecules with unrealistically large carbon rings. Therefore, they added an additional ``$\mbox{RingPenalty}$'' term to penalize rings with more than six carbons. In the ORGANIC GAN code, S\'{a}nchez-Lengling et al.\ added several penalty terms to the original SA score function, and also developed an additional reward for chemical symmetry, based on the observation that symmetric molecules are typically easier to synthesize.\cite{Sanchez-Lengeling2017ORGANICchemrxiv}

For drug molecules, the use of scoring functions developed to estimate how ``drug-like'' or ``natural'' a compound is can also help improve the synthesizability, stability, and usefulness of the generated molecules.\cite{Guimaraes2017ORGANarxiv} Examples of such functions include Lipinski's Rule of Five score,\cite{Lipinski1997:3} the natural product-likeness score,\cite{Ertl2008:68} the quantitative estimate of drug-likeness,\cite{Bickerton2012:90} and the Muegge metrics.\cite{Muegge2001,Muegge2003:302} Another option of particular utility to drug discovery is to apply medicinal chemistry filters either during training or post-training to tag unstable, toxic, or unsynthesizable molecules. For drug molecules, catalogs of problematic functional groups to avoid have been developed in order to limit the probability of unwanted side-effects.\cite{Kalgutkar2005} For energetic molecules and other niche domains an analogous set of functional groups to avoid has yet to be developed.

Many software packages exist for checking molecule's stability and syntheszability which may be integrated into training or as a post-training filter. For example Popova et al.\ use the ChemAxon structure checker software to do a validity check on the generated molecules.\cite{Popova2018} Bjerrum et al.\ use the Wiley ChemPlanner software post-training to find synthesis routes for 25-55\% of the molecules generated by their RNN.\cite{Bjerrum2017arxiv} Finally, Sumita et al.\ check for previously discovered synthetic routes for their generated molecules using a SciFinder literature search.\cite{SumitapreprintACScentral} 

It is worth mentioning that deep learning is now being used for the prediction of chemical reactions\cite{Fooshee2018DeepLearning,Schwaller2018} and synthesis planning. Segler et al.\ trained a deep reinforcement learning system on a large set of reactions that have been published in organic chemistry and demonstrated a system that could predict chemical reactions and plan synthetic routes at the level of expert chemists.\cite{Segler2018Planning}


\subsubsection{Rewards for good properties} 
Because they are called often during training, reward functions should be quick to compute, and therefore fast property estimators are called for. Examples of property estimation functions which are fast to evaluate are the estimated octanol-water partition coefficient (LogP), and the quantitative measure of drug-likeness (QED),\cite{Bickerton2012:90} both of which can be found in the open source package RDKit.\cite{rdkit}

Since physics based prediction is usually very computationally intensive, a popular approach is to train a property predicting machine learning model ahead of time. There is now an enormous literature on deep learning for property prediction demonstrating success in multiple areas.\cite{Butler2018MLforMolecules} Impressive results have been obtained, such as systems which can predict molecular energies at DFT accuracy,\cite{Faber2017JCTC} and highly accurate systems which can transfer between many types of molecules.\cite{cheng2019universal} While the literature on quantum property prediction is perhaps the most developed, success has been seen in other areas, such as calculating high level properties of energetic molecular crystals (such as sensitivity and detonation velocity).\cite{Barnes2018arxiv,Elton2018scirep} Many predictive models are now published for ``off the shelf'' use, for instance a collection of predictive models for ADME (absorption, distribution, metabolism, and excretion) called ``SwissADME'' was recently  published.\cite{Daina2017SwissADME} 

It has also been demonstrated that traditional physics-based simulations can be used\textemdash Sumita et al.\ optimize absorption wavelength by converting SMILES strings into 3D structures using RDKit and then calculating their UV-VIS absorption spectra on-the-fly with time-dependent density functional theory (TD-DFT). Instead of the obvious reward function $-\alpha | \lambda^* - \lambda |$, where $\lambda^*$ is the target wavelength, they used the following:\cite{SumitapreprintACScentral}
\begin{equation}\label{sumitareward}
R  =
    \begin{cases}
     \frac{-\alpha | \lambda^* - \lambda | }{ 1 + \alpha|\lambda^* - \lambda |} & \mbox{If DFT calculation successful} \\
    -1 & \mbox{If DFT calculation fails}
    \end{cases} 
\end{equation}
From the molecules generated by their RNN, Sumita et al.\ selected six molecules for synthesis and found that 5/6 exhibited the desired absorption profiles.\cite{SumitapreprintACScentral}  

A reward function which has been used by several different researchers to generate drug molecules is:   
\begin{equation}\label{popularJscore}
    J(S) = \mbox{logP}(S) - \mbox{SA}(S) - \mbox{RingPenalty}(S) 
\end{equation}

Yang et al.\ add an additional penalty for generating invalid SMILES which could be used more broadly:\cite{Yang2017:972}
\begin{equation}
    R(S) = \begin{cases}
                \frac{J(S)}{1 + |J(S)|} & \mbox{for valid SMILES} \\
                - 1.0 & \mbox{for invalid SMILES} 
            \end{cases}
\end{equation}

In the context of training the ORGAN architecuture, Guimares et al.\ found that rotating the reward function metric from epoch to epoch had some advantages to using all metrics at once.\cite{Guimaraes2017ORGANarxiv} In other words, in one epoch the rewards may just be for diversity, while in the next they would just be for synthesizability, and so on. This idea could likely be explored further.

\section{Prospective and future directions}
In this review we have tried to summarize the current state of the art for generative modeling of molecules using deep learning. The current literature is composed of a rich array of representation strategies and model architectures. As in many areas of generative modeling and deep learning, the present day work is largely empirical in nature. As our mathematical understanding of the landscape of generative models improves, so too will our ability to select the best approaches to a particular problem. There are many promising new techniques and architectures from deep generative modeling and optimization more broadly which are ripe to be transferred to the world of molecules. For example, for sequence modeling the Maximum Likelihood Augmented Discrete GAN (MaliGAN) has been shown to be superior to the SeqGAN on which ORGAN is based.\cite{che2017maximumlikelihood} With RNNs, recently developed attention mechanisms and external memory cells offer a possible avenue to improve SMILES string generation.\cite{Olah2016attention}

It is worth noting that the latest genetic algorithm based methods can still compete with today's deep learning based methods. Yoshikawa et al.\ developed a genetic algorithm which makes edits to SMILES and uses population-based evolutionary selection to generate molecules with high binding affinity as calculated via docking (rDock).\cite{Yoshikawa2018Population} They compared their method with three other deep-learning based methods (CVAE\cite{GmezBombarelli2018:268}, GVAE\cite{KusnerArxiv2017}, and ChemTS\cite{Yang2017:972}) for optimizing the ``penalized log P score'' (eqn.\ \ref{popularJscore}). They found that with computer time fixed to eight hours, their method performed better or comparable to the deep learning methods. In a similar vein, Jensen found that a genetic algorithm performed better than a SMILES based RNN+Bayesian optimization, the ChemTS RNN, and a CVAE with 100x lower computational cost.\cite{Jensen2019GraphBasedGA} It appears that genetic algorithms can explore chemical space in a very efficient manner. 


In our discussion of GANs we highlighted an important way in which GANs are superior to maximum likelihood based methods\textemdash namely that they can avoid the ``filling in'' problem which occurs with maximum likelihood where the model's distribution ends up non-zero where the data distribution is zero. Another point is that the theorems on which the maximum likelihood methodology is based only hold true in the limit of infinite samples.\cite{Boukouvalas2018ThesisArXiv} In general it appears that GANs can be trained with far fewer samples than maximum likelihood based methods\textemdash this can be seen by looking at the $N_{\ff{train}}$ values in table \ref{generativemodelingpapers}. In addition to their benefits, we also touched on several difficulties with GANs\textemdash small starting gradient, training instabilities, the perfect discriminator problem, and mode collapse. However, we also cited possible remedies for each of these issues and we expect more remedies to be developed in the future. 

There are several major trends we have observed which present day practitioners and those entering the field should be cognizant of: 

\textbf{New representation methods} SMILES based techniques are quickly being replaced with techniques that work directly with chemical graphs and three dimensional chemical structures. Directly working with chemical graphs, either by using chemistry-preserving graph operations or tensor representations avoids the problems aropssociated with requiring deep generative models to learn SMILES syntax. At the same time, there is also growing interest in generative models which can generate 3D equilibrium structures, since in many applications the specific 3D positions of atoms can be important. 

\textbf{Better reward functions} As mentioned earlier, reward function design is a critical component to molecular generation and optimization. We expect future work will use more sophisticated reward functions which combine multiple objectives into a single reward function. Using multiple reward functions and multi-objective reinforcement learning is also a promising approach.\cite{Zhou2018arXiv}  

\textbf{Pure reinforcement learning approaches} The deep reinforcement learning work of Zhou et al.\ demonstrated superior molecular optimization performance when compared with the Junction Tree VAE,\cite{Jin2018arxiv} ORGAN,\cite{Guimaraes2017ORGANarxiv} and Graph Convolutional Policy Network\cite{You2018arXiv} approaches when optimizing the logP and QED metrics.\cite{Zhou2018arXiv} The work of Zhou et al.\ is notable as it is the first to take a pure RL approach with no pretrained generator. We believe much future work in molecular optimization will proceed in this direction since many application areas have limited training data available. 

\textbf{Hierarchical modeling} Hierarchical representation schemes will allow for efficient generative modeling of large molecules (such as proteins\cite{Alley2019biorxiv,NIPS2018_7978}) as well as complex systems such as polymers, metal organic frameworks, and molecular crystals. Generative modeling techniques will also be useful not just for optimizing molecules but also optimizing the structures and systems in which they are embedded. GANs have recently been applied to the generation of crystal structures\cite{Nouira2018crystalGANarXiv} and microstructures.\cite{Li2018microstructure,Singh2018arXiv,Yang2018microstructurearxiv} Hierarchical GANs\cite{chen2018hgan} may be useful for the generation of many-molecule complexes or for the simultaneous optimization of both material and geometry in nanomaterials and metamaterials.

\textbf{Closing the loop} After the synthesis and characterization of new lead compounds the data obtained can be fed back to improve the models used, a process called ``closing the loop''. More work is needed to develop workflows and methods to interface and integrate generative models into laboratory platforms to allow for rapid feedback and cycling. A key challenge is developing useful software and cyberinfrastructure for computational screening and data management.\cite{Hachmann2018Building} The potential for efficiency improvements via automated AI-assisted synthesis planning and ``self-driving'' robotic laboratories is quite profound.\cite{Saikin2018Closedloop,Gromski2019review,Tabor2018naturereviews,Henson2018ACSrobots,Roch2018ChemOS}


\section*{Acknowledgements}
Support for this work is gratefully acknowledged from the U.S.\ Office of Naval Research under grant number N00014-17-1-2108 and from the Energetics Technology Center under project number 2044-001. Partial support is also acknowledged from the Center for Engineering Concepts Development in the Department of Mechanical Engineering at the University of Maryland, College Park. We thank Dr.\ Ruth M. Doherty, Dr.\ William Wilson, and Dr.\ Andrey Gorlin for their input and for proofreading the manuscript.


\bibliography{review_article} 
\bibliographystyle{unsrt} 

\end{document}